\documentclass[pdflatex,sn-nature,iicol]{sn-jnl}

\usepackage{graphicx}
\usepackage{multirow}
\usepackage{amsmath,amssymb,amsfonts}
\usepackage{mathrsfs}
\usepackage[title]{appendix}
\usepackage{textcomp}
\usepackage{manyfoot}
\usepackage{booktabs}
\usepackage{mathtools}
\usepackage{listings}

\usepackage{siunitx}
\usepackage[table]{xcolor}
\usepackage{tabularx}
\usepackage{array}
\usepackage{makecell}
\usepackage{ragged2e}
\usepackage{longtable}
\usepackage{url}

\usepackage{enumitem}
\usepackage{dsfont}
\usepackage{fancyvrb}
\usepackage{tikz}
\usetikzlibrary{positioning,arrows.meta,calc,fit,backgrounds}

\usepackage{placeins}

\usepackage[caption=false]{subfig}

\newcolumntype{Y}{>{\RaggedRight\arraybackslash}X}

\renewcommand{\arraystretch}{1.15}
\theoremstyle{thmstyleone}

\theoremstyle{thmstyletwo}

\theoremstyle{thmstylethree}

\begin{document}

\title{Model-Free Runtime Measurement of Task-Relevant Information}

\author*[1]{\fnm{Dimitris} \sur{Panagopoulos}}\email{d.panagopoulos@cranfield.ac.uk}
\author[1]{\fnm{Adolfo} \sur{Perrusqu\'{i}a}}\email{adolfo.perrusquia-guzman@cranfield.ac.uk}
\author[1]{\fnm{Yang} \sur{Xing}}\email{yang.x@cranfield.ac.uk}
\author[1]{\fnm{Weisi} \sur{Guo}}\email{weisi.guo@cranfield.ac.uk}
\affil*[1]{\orgdiv{Faculty of Engineering and Applied Science}, \orgname{Cranfield University}, \orgaddress{\city{Cranfield}, \postcode{MK43 0AL}, \country{UK}}}

\abstract{Autonomous systems increasingly monitor physical kinematics, environmental states and task completion outcomes. While existing work provides methods to plan information seeking actions and interpret acquired observations, current approaches do not monitor whether the live process of information gathering is functioning properly. Consequently, such systems can remain operationally active without resolving outstanding mission requirements. To address this limitation, we introduce Acquisition-State Telemetry (AST), a model free architecture that evaluates ongoing information acquisition from observable evidence relative to declared task requirements. Specifically, a domain interface maps raw observations into task resolution coordinates and normalized comparative representations, which are processed by a fixed causal monitor. This monitor emits two continuous real time signals that quantify (i) prospective progress across unresolved requirements and (ii) informational friction generated when recurrent evidence yields negligible informational gain. We evaluated a single, unadjusted monitor configuration across eight diverse domains including search and rescue dialogue, spatial grid exploration, hierarchical capability acquisition, continuous robotic manipulation, and probabilistic intent estimation. We show that across these domains the friction channel distinguished unproductive stagnation from purposeful exploration prior to task completion, while remaining muted during post completion activity. Furthermore, in closed loop reinforcement learning experiments under early termination constraints, we demonstrate that penalizing informational friction allowed agents to sustain high mission success, whereas unpenalized policies and baselines with matched operational costs failed to complete required tasks.}


\maketitle

\section{Introduction}
\label{sec:introduction}

\begin{figure*}[!t]
    \centering
    \includegraphics[width=\linewidth]{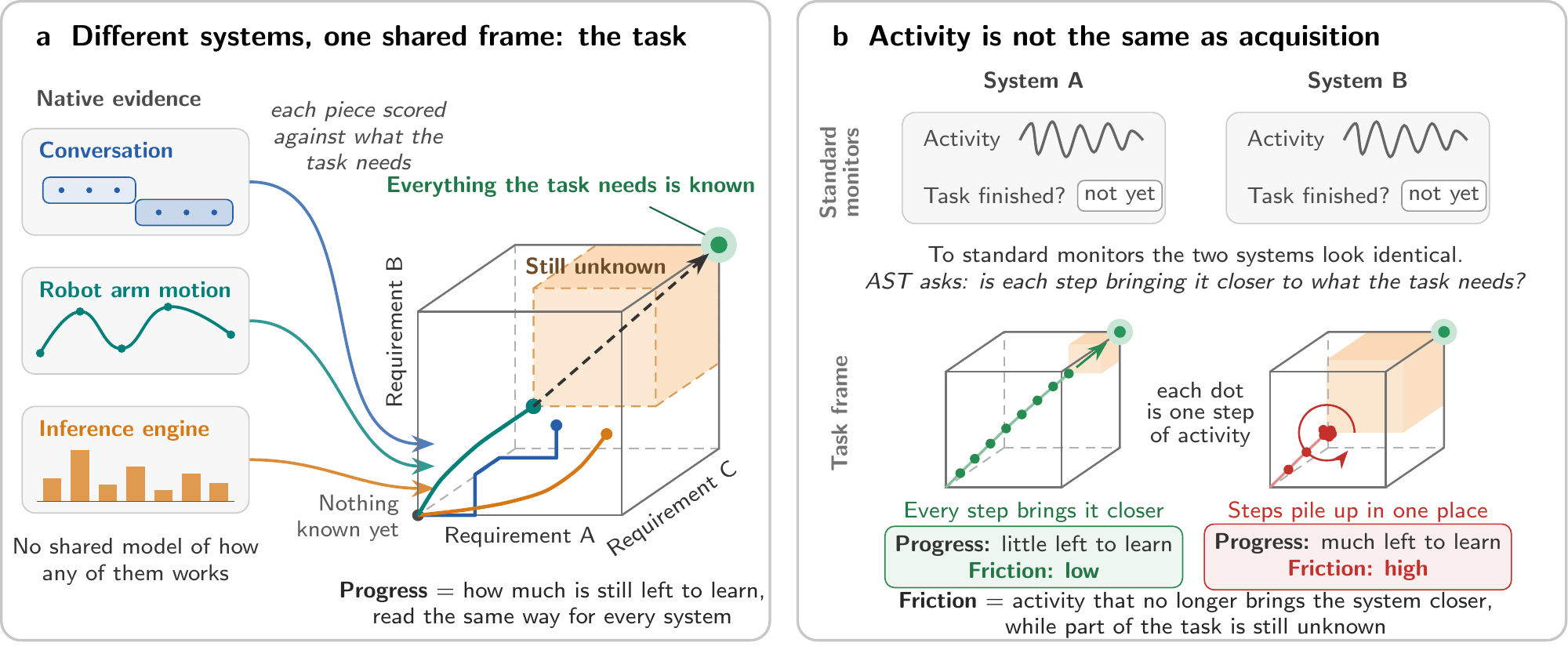}
    \caption{\textbf{Distinguishing operational activity from information acquisition.}
    \textbf{a,} A conversation, a robot arm, and an inference engine generate observations through distinct mechanisms that share no common model, yet all pursue a structured task. Projecting each piece of evidence against task requirements places diverse applications into a shared coordinate frame running from an empty state to complete task resolution. Each system traces a distinct trajectory through this frame, showing stepwise transitions in dialogue, continuous changes in manipulation, and non-linear paths in probabilistic inference, while the remaining unresolved task requirements supply a common frame for comparing their acquisition trajectories.
    \textbf{b,} Two systems exhibiting identical activity levels. Conventional monitors tracking operational liveness and binary task completion cannot distinguish between them prior to termination. Within the task frame, the divergence is immediate, as System A advances steadily toward full resolution with low friction, whereas System B repeatedly cycles through familiar representations while task requirements remain unfulfilled, triggering high friction.}
\label{fig:key_idea}
\end{figure*}

Autonomous cyberphysical systems mostly monitor physical kinematics, environmental surroundings, and binary task completion outcomes \cite{ramesh2022robot}. In particular, sensors track physical trajectories and actuators measure mechanical loads, whereas downstream success metrics record whether an objective was achieved. However, the ongoing process of acquiring the information required to achieve those objectives remains unmonitored at runtime. When operating in dynamic or incompletely modeled environments, an autonomous agent can sustain continuous operational activity while making no material progress toward its mission requirements \cite{ramesh2022robot}. For example, a search and rescue conversational assistant may repeatedly question witnesses without clarifying the coordinates of a missing person \cite{panagopoulos2024selective,gruffeille2024disaster, panagopoulos2026dialogue}. Similarly, a mobile robot may explore a partitioned environment while traversing repetitive loops \cite{chevalier2023minigrid, chiou2021mixed} and a probabilistic estimator may ingest continuous observations without resolving ambiguity among competing hypotheses \cite{zhao2024rescue,higashinaka2016dialogue,panagopoulosBOIR}. Because conventional monitors track operational activity rather than information acquisition, systems displaying identical levels of activity can follow fundamentally different epistemic trajectories (Fig.~\ref{fig:key_idea}).

Current decision-making and diagnostic paradigms do not solve this problem because each addresses a different operational level. Within the observe--orient--decide--act (OODA) framework~\cite{boyd2018discourse}, information acquisition contributes to orientation, where observations acquire meaning relative to operational goals. Methods in active sensing, Bayesian experimental design, and information seeking optimize action selection by predicting future information gain \cite{lindley1956,chaloner1995bayesian,tellex2012toward,deits2013clarifying,rao2018learning,aliannejadi2019asking,lee2018answerer,shukla-etal-2019-ask}. However, these policies assume their forward observation models are dependable and lack mechanisms to verify whether the actual runtime gathering process is yielding progress. Conversely, conversational breakdown detectors and interaction quality estimators assess surface level coherence and retrospective outcomes \cite{deriu2021survey,schmitt2015interaction,ultes2017domain,su2018reward,higashinaka2016dialogue,ng2020improving,terragni2022betold,chen2026failed}, but they rely on domain specific heuristics that cannot generalize across heterogeneous physical or algorithmic systems. Similarly, intrinsic motivation and curiosity driven exploration incorporate novelty bonuses into reinforcement learning \cite{oudeyer2007intrinsic,NIPS2016_afda3322,pathak2017curiosity,houthooft2016vime,graves2017automated,panagopoulos2024selective,pmlr-v119-standley20a,gao2025smartrag}. Although they reward environmental novelty, they do not verify whether that novelty resolves declared mission objectives. This leaves a critical gap that directly motivates this study. Specifically, we lack a real-time, independent monitor that can continuously track whether incoming evidence is actually satisfying task requirements, completely separate from the policy gathering it.
Developing such an independent observer presents a fundamental conceptual challenge. In classical control engineering, separating state estimation from feedback control is achieved through observers such as Kalman and Luenberger filters, which rely on known physical dynamics to infer internal states from sensor measurements \cite{kalman1960new,luenberger1964observing}. However, modern autonomous systems span vastly different domains and share no common physical laws.

To bridge this divide, we introduce Acquisition-State Telemetry (AST), a model-free monitoring framework that adopts declared task requirements, rather than physical dynamics, as the invariant reference frame. By decoupling domain semantics from monitoring logic, AST models interaction as a trajectory through a task-referenced coordinate space. A domain-specific interface projects incoming evidence into cumulative resolution scores across declared requirements alongside normalized representations for temporal comparison. A fixed causal monitor then processes this sequence to emit two continuous runtime channels: (i) prospective progress across unresolved requirements and (ii) informational friction, which detects recurrent evidence when ongoing activity fails to advance open objectives.

We evaluated a single, unretuned monitor configuration across eight diverse experimental substrates spanning human dialogue, spatial grid exploration, capability unlocking, continuous robotic manipulation, and internal Bayesian intent estimation \cite{ye2022multiwoz,anderson1991hcrc,wu2022anno,chevalier2023minigrid,hafner2021crafter,mclean2025metaworld,panagopoulosBOIR}. The experiments establish three primary empirical findings. First, informational friction reliably distinguishes unproductive stagnation from purposeful information-seeking across both linguistic and physical domains, while remaining muted during post-goal actiity. Second, the measurement remains robust when substituting expert human scoring with automated language model ensembles. Third, in closed loop reinforcement learning under early termination constraints, penalizing informational friction enables agents to complete mission requirements, whereas unpenalized policies and operational cost baselines fail due to persistent looping. Together, these results demonstrate that task-referenced acquisition telemetry provides a domain-agnostic foundation for runtime epistemic supervision in autonomous systems.

\section{Acquisition-State Telemetry}
\label{sec:methods_detailed}
\label{sec:method}

\subsection{Task binding and acquisition state}
\label{sec:binding}
\label{sec:state_contract}

Let $\mathcal{M}=\{1,\ldots,N\}$ denote a finite set of declared task requirements and $\mathcal{Y}$ the space of observations produced by the monitored process. A task binding $\mathcal{B}=(\mathcal{M},\mathcal{Y},S,e)$ specifies a resolution map $S:\mathcal{Y}\to[0,1]^N$ and a comparative representation map $e:\mathcal{Y}\to\mathbb{R}^d$. Given evidence $y_t\in\mathcal{Y}$ at exchange $t$, the binding supplies
\begin{equation}
\begin{aligned}
\mathbf{s}(t)&=S(y_t)\in[0,1]^N,\\
\mathbf{r}^{(t)}&=e(y_t),\qquad
\lVert\mathbf{r}^{(t)}\rVert_2=1.
\end{aligned}
\label{eq:dual_projection}
\end{equation}
The coordinate $s_i(t)$ measures the resolution of requirement $i$ supported by the current observation and the unit-norm vector $\mathbf{r}^{(t)}$ provides a geometry for comparing evidence across exchanges. An attribution marker $i_t\in\mathcal{M}\cup\{\bot\}$ identifies the requirement targeted by the preceding action, with $\bot$ denoting an unattributed exchange. Attribution records the target of an attempt and does not constrain which requirements its evidence can resolve.

Assuming that previously verified information remains valid within an episode, the attained-resolution state $\boldsymbol{\upsilon}(t)\in[0,1]^N$ evolves as
\begin{equation}
\begin{aligned}
\upsilon_i(t)&=\max\!\bigl\{\upsilon_i(t-1),s_i(t)\bigr\},\\
\Delta\upsilon_i(t)&=\upsilon_i(t)-\upsilon_i(t-1),
\qquad\upsilon_i(0)=0.
\end{aligned}
\label{eq:state_accumulation}
\end{equation}
Thus $\Delta\upsilon_i(t)\geq0$ is newly attained resolution and $1-\upsilon_i(t)$ is the unresolved fraction of requirement $i$. The signed increment $\delta_i(t)=s_i(t)-\upsilon_i(t-1)$ additionally compares current evidence with the previously attained state and can be negative even though cumulative resolution is non-decreasing. Invalidation of previously verified information ends the episode and requires a reset of the state and evidence histories. The binding and episode conditions are specified in Appendix~\ref{supp_note_formal}.

\begin{figure*}[!t]
    \centering
    \includegraphics[width=\linewidth]{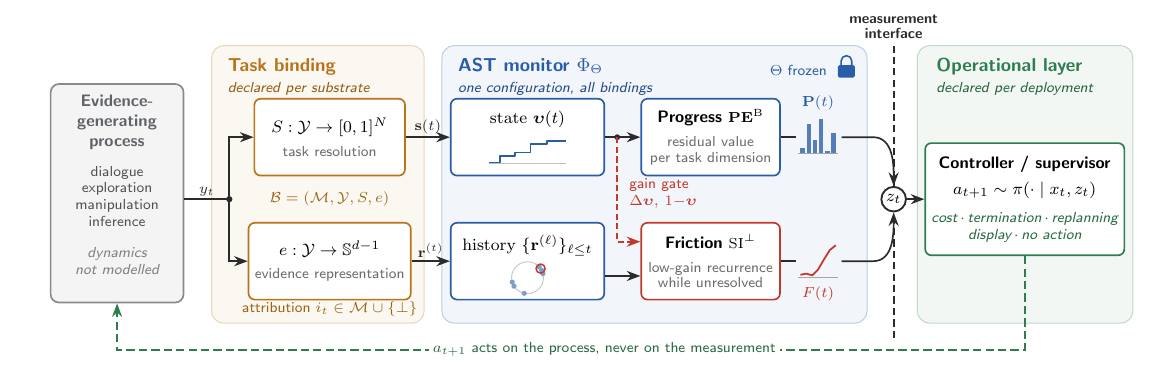}
    \caption{\textbf{The Acquisition-State Telemetry architecture.} \textbf{a,} A task binding maps observations to task-resolution coordinates and unit-norm comparative representations. \textbf{b,} The frozen causal monitor accumulates resolution and reports prospective progress and gain-gated friction. Friction combines representational recurrence, realized gain and residual unresolved volume. \textbf{c,} Downstream policies use telemetry for observation, reward shaping, intervention or termination without changing the monitoring rules.}
    \label{fig:architecture_overview}
\end{figure*}

\subsection{Causal monitoring channels}
\label{sec:channels}

Let $\tau_{1:t}$ denote the ordered evidence and acquisition history available through exchange $t$. The causal monitor $\Phi_{\Theta}$, with fixed numerical configuration $\Theta$, maps this history to two telemetry channels:
\begin{equation}
 z_t=\Phi_{\Theta}(\tau_{1:t})
 =\bigl(\mathbf{P}(t),F(t)\bigr),
 \label{eq:monitor_mapping}
\end{equation}
where $\mathbf{P}(t)\in\mathbb{R}_{\geq0}^N$ estimates prospective acquisition value by requirement and $F(t)\in[0,1]$ measures acquisition friction.

The reference progress channel $\mathbf{PE}^{\mathrm{B}}(t)$ updates a Beta--Bernoulli posterior from attributed attempts and combines estimated attempt yield with resolution-dependent uncertainty, declared prerequisite dependencies and a term based on the straightness of a gain-weighted evidence path. Its expected-information component vanishes as a requirement approaches full resolution. The separate trajectory component depends on directional consistency and can remain non-zero when that requirement is saturated. The full posterior and trajectory updates are given in Appendix~\ref{supp_note_monitor_math}.

For each requirement $i$ with earlier attributed evidence representations, recurrence $\mu_i(t)\in[0,1]$ is the complement of the orthogonal novelty of $\mathbf{r}^{(t)}$ relative to the historical representation subspace. Let $\mathcal{A}_t$ be the set of requirements with such a history. The friction channel has the structural form
\begin{equation}
\begin{aligned}
F(t)&\simeq\sum_{i\in\mathcal{A}_t}
\omega_i(t)\,\mu_i(t)\\
&\qquad\times D\bigl(\varrho(t)\bigr)
\bigl(1-\upsilon_i(t)\bigr),\\
\omega_i(t)&=\frac{\mu_i(t)}
{\sum_{j\in\mathcal{A}_t}\mu_j(t)}.
\end{aligned}
\label{eq:main_friction_structure}
\end{equation}
Here $\omega_i(t)$ weights requirements by their relative recurrence, $D$ gates recurrence according to realized gain $\varrho(t)$, and $1-\upsilon_i(t)$ weights the remaining unresolved requirement. For a directly attributed exchange, $\varrho(t)=\delta_{i_t}(t)$ Unattributed exchanges and bindings with inferred attribution use aggregate gain $\varrho(t)=\sum_i\Delta\upsilon_i(t)$. The evaluated gate $D(\varrho)=\max\{D_{\min},1-\lambda\varrho\}$ attenuates recurrence under positive gain and amplifies it under negative signed gain. Completed requirements contribute zero through the unresolved-volume factor. The recurrence weights apply when total recurrence is positive and friction is zero without an active history or sufficient recurrence. Appendix~\ref{supp_note_monitor_math} specifies the projection, numerical tolerances, regularization and clipping used in the implemented index.

\subsection{Frozen reference implementation and evaluation}
\label{sec:evaluated_monitor}
\label{sec:frozen_monitor}
\label{sec:methods_conventions}

The evaluated implementation instantiates $\mathbf{P}(t)$ as the Bayesian Progress Estimator $\mathbf{PE}^{\mathrm{B}}(t)$ and $F(t)$ as the orthogonal-subspace Stalling Index $\mathrm{SI}^{\perp}(t)$. The monitor configuration $\Theta$ and its update rules remain fixed across eight substrates. Each substrate supplies its task-specific resolution and representation maps. The monitor updates causal statistics without parameter training. Downstream policies are trained separately in the closed-loop reinforcement-learning experiment.

\section{Experiments \& Results}
\label{sec:results}
\label{sec:experiments}
\subsection{Cross-domain evaluation protocol}
\label{sec:setup}

We evaluated the frozen AST monitor across eight benchmarks spanning human--agent dialogue, discrete spatial exploration, capability unlocking, continuous robotic manipulation, and embedded probabilistic inference (Extended Data Table~\ref{tab:eval_map}; binding specifications in Appendix~\ref{supp_note_bindings} and Appendix Table~\ref{tab:supp_bindings}). The evaluation covers transfer of the frozen configuration $\Theta$ between substrate bindings, discrimination of pre-resolution loops and post-completion recurrence, acquisition dynamics under discrete and incremental task structures, and the effect of telemetry on downstream policy learning.

We applied the same diagnostic thresholds across all eight domains. Specifically, exchanges where \(\mathrm{SI}^{\perp}(t) > \theta_{\mathrm{SI}} = 0.20\) are classified as stalled, and coordinates reaching \(\upsilon_i(t) \geq \upsilon_{\mathrm{complete}} = 0.80\) are considered resolved. In environments with controlled loop injections, including MiniGrid~\cite{chevalier2023minigrid}, Crafter~\cite{hafner2021crafter}, and MetaWorld~\cite{mclean2025metaworld}, we quantified event-level stall discrimination using ROC-AUC, precision, recall, and temporal span coverage against ground-truth loop windows. In continuous session benchmarks, such as SAR, MultiWOZ~\cite{ye2022multiwoz}, Map Task~\cite{anderson1991hcrc}, AnnoMI~\cite{wu2022anno}, and BOIR~\cite{panagopoulosBOIR}, we evaluated system-level performance through cumulative knowledge acquisition, session stall burden representing the fraction of exchanges exceeding \(\theta _{\mathrm{SI}}\), and multi-horizon (\(k \in \{1,3,5\}\)) top-1 next-gain prediction accuracy for the progress channel \(\mathbf{PE}^{\mathrm{B}}\).

\subsection{Controlled inquiry regimes}
\label{sec:exp1}

A simulated SAR witness interview requiring eight missing-person categories provided three controlled acquisition regimes: systematic inquiry resolving the requirements (\textbf{S1}), repetitive agent questioning with uninformative witness replies (\textbf{S2}), and repetitive witness replies that leave requirements unresolved (\textbf{S3}). We also repeated the evaluation with automated evidence scoring.

\begin{figure*}[!t]
\centering
\captionsetup{font=small,skip=3pt}
\captionsetup[subfloat]{font=footnotesize,skip=1pt}

\includegraphics[width=\linewidth,height=0.6\textheight,keepaspectratio]{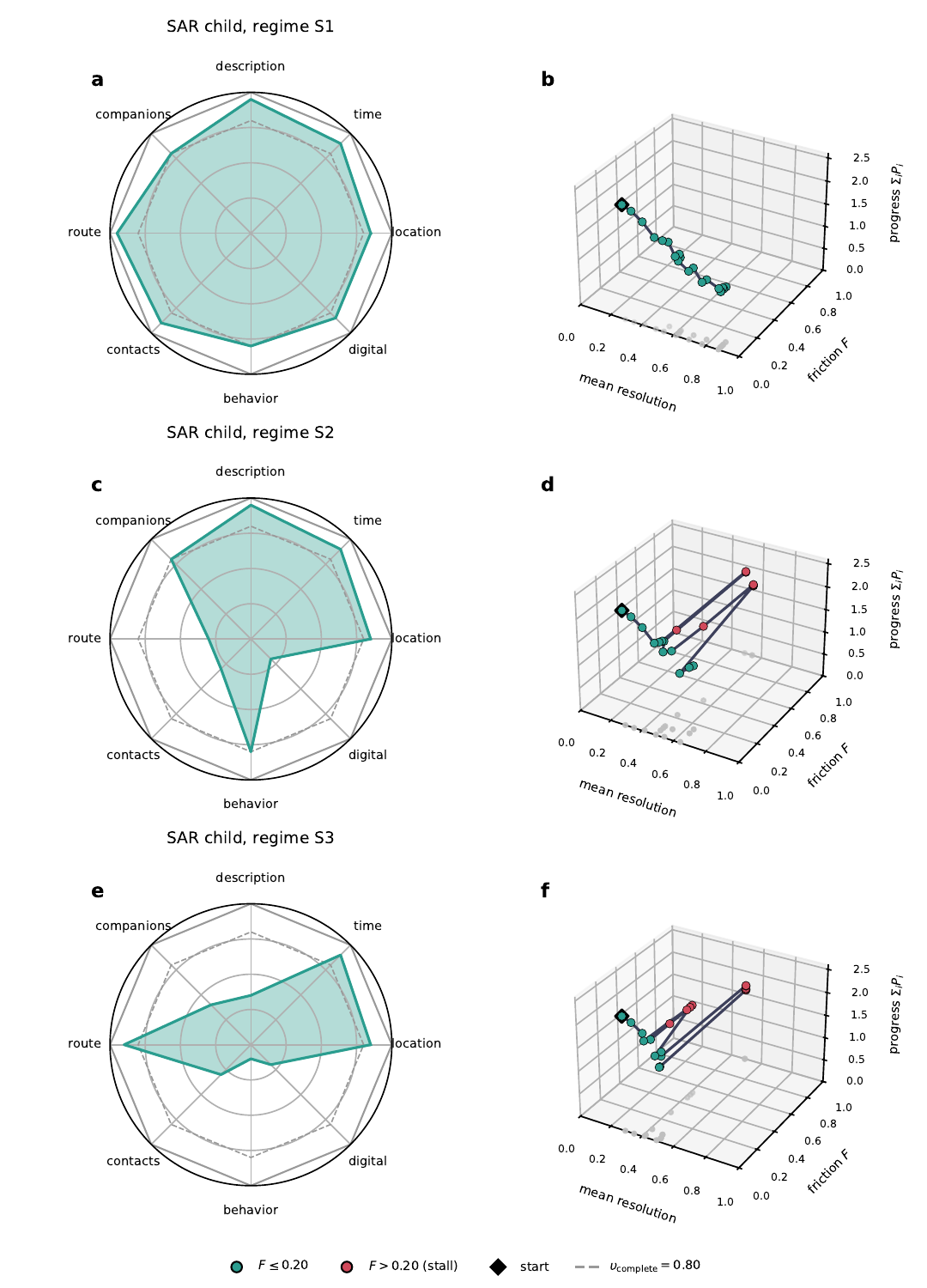}
\caption{\textbf{AST telemetry across controlled SAR inquiry regimes (child corpus, expert scores).} \textbf{a,c,e,} Final resolution of each coordinate (filled) against full resolution (outer ring) and the completion threshold $\upsilon_{\mathrm{complete}}=0.80$ (dashed) for regimes S1, S2 and S3. \textbf{b,d,f,} Corresponding trajectories through mean resolution, friction $\mathrm{SI}^{\perp}(t)$ and total progress $\sum_i\mathrm{PE}^{\mathrm{B}}_i(t)$; each marker is an exchange, red markers exceed $\theta_{\mathrm{SI}}=0.20$, the diamond marks the first exchange and grey points are projections onto the floor. S1 advances to full resolution with negligible friction, whereas S2 and S3 stall before completion with friction excursions close to 1.}
\label{fig:sar_trajectory}
\end{figure*}

The frozen monitor separated all three behavioral regimes (Fig.~\ref{fig:sar_trajectory} and Appendix Table~\ref{tab:supp_sar_scorer}). Efficient inquiry \((\textbf{S1})\) resolved all 8.00 ± 0.00 categories with zero stall burden and near-zero mean friction (0.02 ± 0.004). Both recurrent regimes stopped near half-completion and produced elevated friction. Repetitive questioning \((\textbf{S2})\) incurred a 35.00 ± 7.07\% stall burden with a mean friction of 0.26 ± 0.08 and 4.33 ± 0.94 categories resolved. Witness non-informativeness \((\textbf{S3})\) incurred a 36.67 ± 12.47\% stall burden with a mean friction of 0.28 ± 0.09 and 4.67 ± 1.25 categories resolved. The two stalling regimes produce comparable low-yield acquisition traces despite different sources of repetition, indicating that friction responds to evidence relative to requirements rather than the identity of the source.

We substituted the expert resolution map $S$ with the mean of five local language-model scorers (pooled Krippendorff's \(\alpha\) = 0.80; Appendix~\ref{supp_note_scorer}). Despite yielding lower absolute completion counts, the ordering of friction across regimes remained similar under the frozen monitor. Specifically, friction values scaled from 0.04 ± 0.01 for S1 (3.33\% stall burden), to 0.20 ± 0.05 for S2 (35.00\%) and 0.24 ± 0.11 for S3 (33.33\%; Appendix Table~\ref{tab:supp_sar_scorer}). The scorer substitution preserved the relative friction ranking and changed the absolute telemetry values.

\subsection{Human-interaction benchmarks}
\label{sec:human_boundary}

We evaluated three types of natural human interaction: atomic slot-filling (MultiWOZ~\cite{ye2022multiwoz}), incremental collaborative grounding (HCRC Map Task~\cite{anderson1991hcrc}), and reflective clinical counseling (AnnoMI~\cite{wu2022anno}).

\subsubsection{MultiWOZ: atomic slot filling}
\label{sec:exp2}
In transactional booking dialogues, requirements typically resolve in single-turn discrete steps upon mention. Consequently, 39 of the 40 evaluated MultiWOZ sessions (97.5\%) contained no stalled exchanges. Across multi-slot transitions (\(n=56\)), the progress channel \(\mathbf{PE}^{\mathrm{B}}\) predicted the next resolved booking slot with 51.8\% top-1 accuracy, nearly tripling a uniform guessing baseline (17.9\%, \(p=9.1\times10^{-9}\)), whereas a recency baseline failed completely (0\%) due to the lack of multi-turn incremental accumulation.

\subsubsection{HCRC Map Task: incremental grounding}
\label{sec:exp3}
Collaborative route navigation grounds spatial landmarks incrementally across multiple turns. Across 27 Map Task dialogues, \(\mathbf{PE}^{\mathrm{B}}\) predicted the next grounded landmark with a top-1 accuracy 13.7 percentage points above chance at horizon \(k=5\) (\(p=1.4\times10^{-21}\)), outperforming the baseline in 21 of 27 sessions (sign test \(p=0.006\)). Reflecting the exploratory and clarifying nature of spatial navigation, Map Task incurred a mean stall burden of 8.0\%, contrasting with the near-zero baseline friction of atomic slot-filling.

\subsubsection{AnnoMI: reflective clinical interaction}
\label{sec:exp4}
In 20 motivational interviewing sessions mapped across five client-state dimensions representing the presenting problem, motivation, barriers, resources, and action readiness, AST friction tracked informational throughput (Appendix~\ref{supp_note_human} and Appendix Table~\ref{tab:supp_annomi}). Sessions with higher stall burden yielded significantly lower information gain per turn (Spearman \(\rho=-0.666\), \(p=0.0014\)) and lower cumulative knowledge resolution (\(\rho=-0.570\), \(p=0.0087\)). Notably, stall burdens were similar for sessions rated as high- and low-quality counseling by external clinicians (35.20\% and 35.77\%, respectively; \(p=0.910\)). This shows the scope of the AST reporting acquisition relative to the five declared client-state dimensions and not clinical quality, which remains a separate outcome as reflective pauses and empathetic pacing are common in counseling and have been discussed as positive friction in conversational artificial intelligence~\cite{inan2025better}.

\subsection{Robotic and inference benchmarks}
\label{sec:engine_bindings}

We applied the same frozen monitor to four additional (non-linguistic) systems: 2D spatial exploration (MiniGrid~\cite{chevalier2023minigrid}), hierarchical capability unlocking (Crafter~\cite{hafner2021crafter}), continuous kinematic manipulation (MetaWorld~\cite{mclean2025metaworld}), and embedded Bayesian intent estimation (BOIR~\cite{panagopoulosBOIR}).

\subsubsection{MiniGrid: incremental spatial exploration} \label{sec:minigrid}
In a gridworld partitioned into 3\(\times\)3 spatial sectors where resolution tracks the fraction of revealed walkable tiles, AST detected injected movement loops inside incomplete sectors under pre-resolution stalling, designated as Condition I1. For this condition, performance reached an ROC-AUC of 0.882, precision of 0.742, recall of 0.745, and span coverage of 0.800 (Extended Data Table~\ref{tab:nonling_si}), with mean friction rising from 0.103 during nominal exploration to 0.334 during stalls. Conversely, when identical movement loops occurred inside already explored sectors under post-goal waiting, designated as Condition I2, the unresolved volume gate \((1-\upsilon_i(t))\) attenuated the signal, reducing mean friction to 0.179 with an ROC-AUC of 0.715.

\subsubsection{Crafter: discrete achievement unlocks}
\label{sec:crafter}
Across Crafter's 22-node achievement tree, $95.7\%$ of positive updates occur as discrete milestones. When agents encountered injected no-op loops prior to unlocking prerequisites, mean friction rose from a nominal baseline of $0.133$ to $0.509$ (ROC-AUC $0.767$, span coverage $0.952$; Extended Data Table~\ref{tab:nonling_si}). Furthermore, because the resolution map scores each achievement by the fraction of its prerequisites satisfied, $\mathbf{PE}^{\mathrm{B}}$ predicted upcoming unlocks with $0.544$ top-1 accuracy at horizon $k=5$, more than doubling both recency ($0.228$) and uniform ($0.211$) baselines (Appendix Table~\ref{tab:supp_progress_acc}).

\subsubsection{MetaWorld: continuous robotic manipulation}
\label{sec:metaworld}
\begin{figure*}[!t]
\centering
\captionsetup{font=small,skip=3pt}
\captionsetup[subfloat]{font=footnotesize,skip=1pt}

\includegraphics[width=\linewidth,height=0.5\textheight,keepaspectratio]{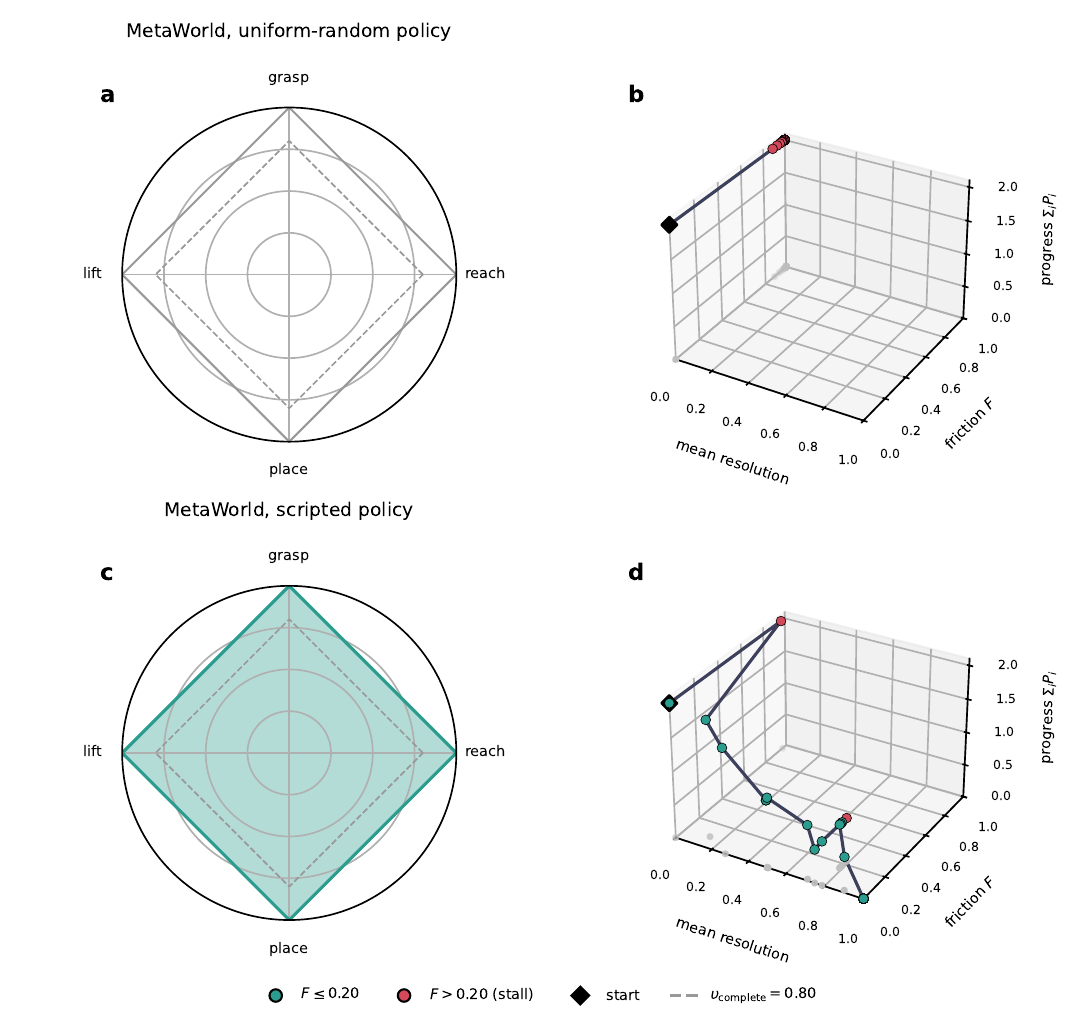}
\caption{\textbf{AST telemetry on MetaWorld pick-and-place (seed 0).} \textbf{a,b,} A uniform-random policy resolves none of the reach, grasp, lift and place milestones, and friction saturates near 1 from the second exchange onwards. \textbf{c,d,} The scripted controller resolves the four milestones in sequence; friction exceeds $\theta_{\mathrm{SI}}$ only at the second exchange, before the first milestone is reached, and once briefly ($0.22$) before placement. Panels and markers as in Fig.~\ref{fig:sar_trajectory}.}
\label{fig:metaworld_trajectory}
\end{figure*}

In continuous pick-and-place manipulation tracked across four kinematic milestones (reach, grasp, lift, and place; Fig.~\ref{fig:metaworld_trajectory}), the gain gate differentiated the two controlled recurrence conditions. When the end-effector entered repetitive orbiting loops before securing the object (pre-resolution stall, Condition I1), friction increased from $0.051$ to $0.999$, with ROC-AUC $0.983$, recall $1.000$, span coverage $1.000$ (Extended Data Table~\ref{tab:nonling_si}).

In contrast, when the arm executed repetitive holding motions at the target location \emph{after} completing the placement milestone (post-goal station-keeping, Condition I2), saturation of the acquisition state $\boldsymbol{\upsilon}(t)$ silenced the friction channel ($0.039$ event mean versus $0.041$ baseline), yielding chance-level discrimination (ROC-AUC $0.499$). Across the two manipulation conditions, the gain gate strongly separated pre-grasp stalling from post-goal holding. Concurrently, $\mathbf{PE}^{\mathrm{B}}$ achieved a top-1 next-milestone prediction accuracy of $0.985$ at $k=5$ (Appendix Table~\ref{tab:supp_progress_acc}).

\subsubsection{BOIR: embedded intent inference}
\label{sec:boir_estimator}
We also applied AST to an internal Bayesian intent estimator inferring human operator goals from continuous trajectories~\cite{panagopoulosBOIR}. In scenario S3, the estimator began all $60$ epochs with an incorrect maximum a posteriori hypothesis (posterior confidence above $0.70$ in $42$ epochs). The monitored true-hypothesis resolution rose from $0.142$ to $0.966$ (Appendix~\ref{supp_note_nonling} and Appendix Table~\ref{tab:supp_boir}). Friction remained low immediately following goal switches ($0.014$--$0.075$) and rose mid-epoch ($0.051$--$0.130$) as competing evidence accumulated during hypothesis re-adjudication.

\subsection{Closed-loop control with epistemic feedback}
\label{sec:exp5}

To evaluate downstream use of acquisition telemetry, we trained five Proximal Policy Optimization (PPO) variants on three SAR corpora (10 random seeds per variant and condition; Appendix~\ref{supp_note_rl}): an uninstrumented baseline (\textbf{Baseline}); an agent paying an operational-cost penalty for repeated actions and exchanges without realized gain, calibrated to match the mean magnitude of the friction penalty (\textbf{Coupled-Blind}); an agent receiving AST telemetry within its observation vector without reward shaping (\textbf{AST-Obs}); an agent observing progress while penalizing AST friction in the reward (\textbf{AST-Pen}); and an agent receiving full telemetry in both observations and reward shaping (\textbf{Full-AST}).

\begin{figure*}[!t]
    \centering
    \captionsetup{font=small,skip=3pt}
    \captionsetup[subfloat]{font=footnotesize,skip=1pt}

    \includegraphics[width=\linewidth,height=0.5\textheight,keepaspectratio]{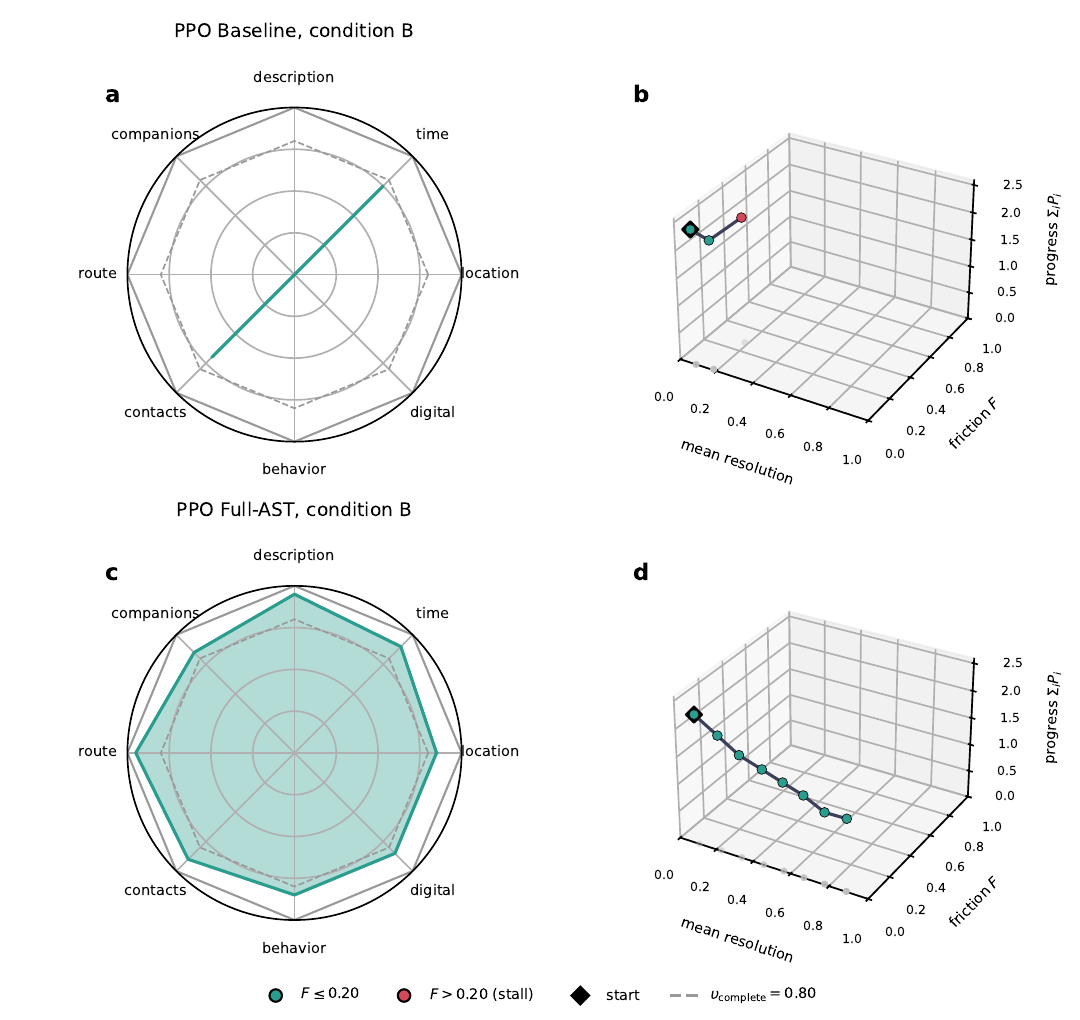}
    \caption{\textbf{Closed-loop PPO episodes under friction-triggered early stopping (Condition B; child corpus, seed 0).} \textbf{a,b,} The Baseline policy ends the episode at the third exchange, when friction exceeds $\theta_{\mathrm{SI}}$, with no category resolved. \textbf{c,d,} The Full-AST policy resolves all eight categories in eight exchanges without exceeding $\theta_{\mathrm{SI}}$. Panels and markers as in Fig.~\ref{fig:sar_trajectory}.}
\label{fig:RL_viz}
\end{figure*}

All policies were evaluated under two operational regimes, including a standard horizon where episodes run until an interaction budget expires, designated as Condition A, and a friction-sensitive early-stopping regime that terminates an episode immediately if \(\mathrm{SI}^{\perp}(t) > \theta_{\mathrm{SI}} = 0.20\), designated as Condition B. Condition B represents an interaction setting with an externally enforced limit on low-yield exchanges. Under an unconstrained interaction budget within Condition A, all policies resolved between 7.67 and 8.00 of the eight mission categories (Table~\ref{tab:rl_closed_loop_eval}). However, under early stopping within Condition B, performance diverged (Fig.~\ref{fig:RL_viz}). The unpenalized Baseline frequently entered repetitive questioning loops that triggered early termination, resolving only 2.37 categories, which represents a 7\% full-resolution rate and a 26.53\% stall burden. The AST-Obs variant, which received telemetry only in observations, resolved 1.73 categories, with a 10\% full-resolution rate and a 37.47\% stall burden.

The matched operational cost penalty, designated as Coupled-Blind, improved performance over the baseline but resolved only 3.53 categories, which corresponds to a 20\% full-resolution rate and a 12.90\% stall burden. This penalty applies to repeated actions and exchanges without immediate gain, including productive follow-up queries. The friction-penalized policies recorded higher completion under early stopping. Specifically, AST-Pen resolved 7.07 of eight categories, representing an 87\% full-resolution rate, and Full-AST resolved 6.93 categories, representing an 83\% full-resolution rate. Both approaches recorded zero stall burden in this evaluation.

\begin{table*}[t]
\centering
\setlength{\tabcolsep}{4pt}
\caption{\textbf{Closed-loop PPO performance under budget termination (Condition A) and friction-sensitive early stopping (Condition B).} Values report means and 95\% bootstrap confidence intervals over 10 random seeds, each averaged over the three SAR corpora (one deterministic evaluation episode per policy). Bold indicates top performance under Condition B.}
\label{tab:rl_closed_loop_eval}
\footnotesize
\resizebox{\textwidth}{!}{%
\begin{tabular}{clrrrrr}
\toprule
\textbf{Cond.} & \textbf{Policy variant} & \makecell{\textbf{Categories}\\\textbf{resolved}} & \makecell{\textbf{Total}\\\textbf{knowledge}} & \makecell{\textbf{Full-res.}\\\textbf{rate}} & \makecell{\textbf{Mean}\\\textbf{friction ($\mathrm{SI}^{\perp}$)}} & \makecell{\textbf{Stall}\\\textbf{burden (\%)}} \\
\midrule
\multirow{5}{*}{A} 
 & Baseline     & 7.67 [7.00, 8.00] & 0.840 [0.766, 0.878] & 0.93 [0.80, 1.00] & 0.009 [0.002, 0.022] & 0.00 [0.00, 0.00] \\
 & Coupled-Blind& 8.00 [8.00, 8.00] & 0.874 [0.872, 0.877] & 1.00 [1.00, 1.00] & 0.004 [0.003, 0.005] & 0.00 [0.00, 0.00] \\
 & AST-Obs      & 7.80 [7.40, 8.00] & 0.854 [0.808, 0.879] & 0.97 [0.90, 1.00] & 0.007 [0.003, 0.016] & 0.00 [0.00, 0.00] \\
 & Full-AST     & 8.00 [8.00, 8.00] & 0.887 [0.885, 0.890] & 1.00 [1.00, 1.00] & 0.002 [0.002, 0.002] & 0.00 [0.00, 0.00] \\
 & AST-Pen      & 8.00 [8.00, 8.00] & 0.887 [0.884, 0.890] & 1.00 [1.00, 1.00] & 0.002 [0.002, 0.002] & 0.00 [0.00, 0.00] \\
\midrule
\multirow{5}{*}{B} 
 & Baseline     & 2.37 [1.67, 3.10] & 0.348 [0.275, 0.426] & 0.07 [0.00, 0.17] & 0.119 [0.101, 0.132] & 26.53 [21.34, 31.98] \\
 & Coupled-Blind& 3.53 [2.67, 4.37] & 0.515 [0.431, 0.596] & 0.20 [0.10, 0.30] & 0.065 [0.048, 0.084] & 12.90 [8.86, 17.43] \\
 & AST-Obs      & 1.73 [0.73, 3.03] & 0.266 [0.151, 0.398] & 0.10 [0.00, 0.27] & 0.367 [0.279, 0.442] & 37.47 [27.58, 45.78] \\
 & Full-AST     & 6.93 [6.23, 7.60] & 0.772 [0.698, 0.844] & 0.83 [0.73, 0.93] & 0.005 [0.002, 0.009] & \textbf{0.00 [0.00, 0.00]} \\
 & AST-Pen      & \textbf{7.07 [6.33, 7.77]} & \textbf{0.787 [0.709, 0.861]} & \textbf{0.87 [0.77, 0.97]} & \textbf{0.004 [0.002, 0.007]} & \textbf{0.00 [0.00, 0.00]} \\
\bottomrule
\end{tabular}%
}
\end{table*}

\section{Discussion}
\label{sec:discussion}

Acquisition-State Telemetry demonstrates that the runtime health of an information gathering process can be tracked independently of the operational policy generating the evidence. By grounding the observer in declared mission requirements rather than inventing physical equations of motion, this architecture establishes a common measurement interface across heterogeneous computational and cyberphysical systems. This decoupling mirrors the foundational separation of state estimation and feedback control in classical systems engineering. Domain specific interfaces translate distinct forms of evidence into shared coordinates of cumulative resolution and comparative representation, allowing a single fixed causal monitor to evaluate epistemic progress across linguistic, robotic, and algorithmic environments.

The selective behavior of informational friction clarifies why traditional repetition penalties and loop detectors fall short in complex operational environments. Repetitive activity is not uniformly harmful. For example, an agent may linger to stabilize a physical grasp or rephrase a statement to facilitate mutual understanding. Our gain-gated friction telemtry signal accounts for this context by treating repetition as detrimental only when it coincides with stagnant task resolution. When incoming evidence yields new informational gains or when targeted objectives reach completion the friction signal attenuates. This mechanism explains why the monitor distinguishes unproductive pre goal looping from legitimate post goal activity across both spatial exploration and robotic manipulation benchmarks. In particular, the monitor isolates those exchanges that consume resources without advancing declared requirements instead of penalizing operational activity indiscriminately.

Moreover, these findings show the relationship between epistemic throughput and broader interaction dynamics. In transactional exchanges and discrete capability unlocking, information arrives in decisive milestones, resulting in minimal baseline friction. In contrast, collaborative grounding and reflective counseling naturally produce sustained recurrence as participants iteratively negotiate meaning or build interpersonal trust. The observed negative correlation between friction and information acquisition confirms that the telemetry accurately reflects operational stagnation relative to declared task coordinates. Crucially, however, identical friction levels can occur across interactions rated high and low in interpersonal quality. This distinction emphasizes that informational friction measures epistemic throughput rather than conversational rapport or therapeutic efficacy. Pauses, reflective summaries, and deliberate pacing often serve essential social or cognitive functions. Runtime telemetry provides an objective measurement of task progress, leaving the decision of how to balance throughput against social rapport to domain specific controllers.

In closed loop decision making, the results demonstrate that passive telemetry observation is insufficient to resolve persistent failure modes. Agents that observe progress signals or get penalized for operational repetition continue to cycle especially if they run out of time. Operational penalties inevitably punish necessary exploratory actions along with redundant ones. By contrast, incorporating gain-gated friction into reward shaping provides a targeted epistemic supervisory signal. Policies trained under this feedback learn to abandon unyielding inquiries and redirect operational effort toward alternative unresolved requirements. Under thgis controlled setting, this capability establishes that runtime acquisition monitoring could also be used as an actionable supervisory signal that can guide search, active sensing, and escalation policies in resource-constrained deployments.

The scope and validity of this monitoring approach depend on specific structural boundaries. The architecture presumes a finite, declared schema whose requirements retain consistent definitions throughout an operational episode. Monotonic accumulation assumes that previously verified information remains valid until an explicit episode reset occurs. Consequently, operational settings involving transient observations, time-varying facts, or active misinformation require extending the state contract beyond monotonic updates. Furthermore, the absolute magnitude of emitted telemetry remains dependent on the resolution mapping and representation spaces provided by the host system. Extending this framework to accommodate dynamically evolving or learnable task schemas, asynchronous multimodal observation streams and automated belief revision represents a critical direction for future investigation. Within these operational boundaries, AST provides a reproducible, model free interface that establishes whether ongoing activity is successfully resolving mission objectives.

\section{Conclusion}
In this work, we introduced Acquisition-State Telemetry, a task-referenced framework for measuring information acquisition through a fixed causal monitor and task-specific evidence bindings. Across eight heterogeneous substrates, its progress and friction channels characterized acquisition dynamics, while friction-based reward shaping improved requirement completion in the evaluated closed-loop setting. Thus, AST provides a common runtime interface for tracking how ongoing activity advances declared information requirements.

\section*{Code availability}
The AST monitor, substrate bindings and experiment scripts, with a single command that reproduces every result, table and figure, are available at \url{https://github.com/dimipan/ast-telemetry/}. MultiWOZ 2.4~\cite{ye2022multiwoz}, the HCRC Map Task corpus~\cite{anderson1991hcrc} and AnnoMI~\cite{wu2022anno} are available from their original releases; the derived session files used here are included.

\bibliography{sn-bibliography}

\clearpage
\setcounter{table}{0}
\makeatletter
\@ifundefined{theHtable}{}{\renewcommand{\theHtable}{ed.table.\arabic{table}}}
\makeatother
\captionsetup[table]{name=Extended Data Table}

\begin{table*}[!t]
\centering
\caption{\textbf{Evaluation map across eight experimental substrates.} A single frozen monitor configuration $\Phi_\Theta$ is evaluated across linguistic, symbolic, continuous robotic, and algorithmic bindings.}
\label{tab:eval_map}

\footnotesize
\setlength{\tabcolsep}{4pt}
\renewcommand{\arraystretch}{1.15}

\begin{tabularx}{\textwidth}{@{}
  >{\RaggedRight\arraybackslash}p{0.14\textwidth}
  >{\RaggedRight\arraybackslash}p{0.16\textwidth}
  Y
  Y
  >{\RaggedRight\arraybackslash}p{0.12\textwidth}
@{}}
\toprule
\textbf{Study} &
\textbf{Substrate} &
\textbf{Acquisition structure} &
\textbf{Primary evidence} &
\textbf{Focus} \\
\midrule

1 --- Controlled anchor &
SAR &
Scripted efficient and recurrent questioning regimes &
Monitor sensitivity to controlled behavioral regimes; invariance to automated scorer substitution &
Cross-domain reuse \\

\addlinespace[2pt]

2 --- Human interaction &
MultiWOZ~\cite{ye2022multiwoz} &
Predominantly atomic slot filling &
Next-slot progress prediction under minimal baseline friction &
Reuse; task structure \\

&
Map Task~\cite{anderson1991hcrc} &
Incremental collaborative grounding &
Multi-horizon landmark prediction and sustained friction during iterative clarification &
Reuse; task structure \\

&
AnnoMI~\cite{wu2022anno} &
Reflective motivational interviewing &
Friction correlation with epistemic throughput and independence from clinical rapport ratings &
Reuse; task structure \\

\addlinespace[2pt]

3 --- Non-linguistic &
MiniGrid~\cite{chevalier2023minigrid} &
Incremental 2D spatial exploration &
Discrimination of trajectory loops in incomplete sectors versus post-goal loitering &
Reuse; friction selectivity \\

&
Crafter~\cite{hafner2021crafter} &
Discrete achievement unlock tree &
Detection of injected no-op loops and prerequisite-guided unlock forecasting &
Reuse; task structure \\

&
MetaWorld~\cite{mclean2025metaworld} &
Continuous robotic manipulation &
Discrimination of pre-grasp orbiting stalls versus post-completion goal holding &
Reuse; friction selectivity \\

&
BOIR~\cite{panagopoulosBOIR} &
Embedded Bayesian intent estimation &
Runtime tracking of internal probabilistic hypothesis adjudication &
Cross-domain reuse \\

\addlinespace[2pt]

4 --- Control &
SAR (PPO) &
Budget-constrained and early-stopped inquiry &
Ablation of passive telemetry observation against gain-gated friction and operational-cost reward penalties &
Measurement--control separation \\

\bottomrule
\end{tabularx}
\end{table*}

\begin{table*}[!t]
\centering
\caption{\textbf{Study 3: Stalling Index ($\mathrm{SI}^{\perp}$) event discrimination across non-linguistic bindings ($\theta_{\mathrm{SI}}=0.20$).} Condition I1 evaluates pre-resolution stalling (repetitive loops while task volume remains unresolved). Condition I2 evaluates post-goal loitering (repetition after task completion). $\mathrm{SI}_{\text{event}}$ and $\mathrm{SI}_{\text{clean}}$ report mean friction inside injected loops and nominal execution windows, respectively.}
\label{tab:nonling_si}

\footnotesize
\setlength{\tabcolsep}{6pt}
\renewcommand{\arraystretch}{1.12}

\begin{tabularx}{\textwidth}{@{}>{\RaggedRight\arraybackslash}Xcccccc@{}}
\toprule
\textbf{Condition} &
\textbf{AUC} &
\makecell{\textbf{Precision}\\\textbf{@$\theta$}} &
\makecell{\textbf{Recall}\\\textbf{@$\theta$}} &
\makecell{\textbf{Span}\\\textbf{coverage}} &
\textbf{$\mathrm{SI}_{\text{event}}$} &
\textbf{$\mathrm{SI}_{\text{clean}}$} \\
\midrule
MiniGrid I1 (residual stall)       & 0.882 & 0.742 & 0.745 & 0.800 & 0.334 & 0.103 \\
MiniGrid I2 (post-goal loiter)     & 0.715 & 0.455 & 0.467 & 0.500 & 0.179 & 0.106 \\
Crafter (injected no-op loop)      & 0.767 & 0.853 & 0.698 & 0.952 & 0.509 & 0.133 \\
MetaWorld I1 (pre-grasp orbit)     & 0.983 & 0.800 & 1.000 & 1.000 & 0.999 & 0.051 \\
MetaWorld I2 (post-goal holding)   & 0.499 & 0.213 & 0.036 & 0.050 & 0.039 & 0.041 \\
\bottomrule
\end{tabularx}
\end{table*}

\clearpage
\onecolumn
\appendix
\numberwithin{equation}{section}
\numberwithin{table}{section}
\numberwithin{figure}{section}

\makeatletter
\providecommand{\theHsection}{}\providecommand{\theHsubsection}{}
\providecommand{\theHequation}{}\providecommand{\theHtable}{}\providecommand{\theHfigure}{}
\renewcommand{\theHsection}{appendix.section.\Alph{section}}
\renewcommand{\theHsubsection}{appendix.subsection.\Alph{section}.\arabic{subsection}}
\renewcommand{\theHequation}{appendix.equation.\Alph{section}.\arabic{equation}}
\renewcommand{\theHtable}{appendix.table.\Alph{section}.\arabic{table}}
\renewcommand{\theHfigure}{appendix.figure.\Alph{section}.\arabic{figure}}
\makeatother
\captionsetup[table]{name=Appendix Table}
\captionsetup[figure]{name=Appendix Fig.}

\section{Substrate Binding Instantiations and Domain Schemas}
\label{supp_note_bindings}

Each evaluated substrate implements the task binding $\mathcal{B}=(\mathcal{M},\mathcal{Y},S,e)$ defined in Section~\ref{sec:binding}. The set $\mathcal{M}$ specifies the declared task requirements and $\mathcal{Y}$ specifies the observation space. The map $S(y_t)$ yields task-resolution coordinates and $e(y_t)$ yields a unit-norm representation for temporal comparison.

Appendix Table~\ref{tab:supp_bindings} specifies the eight substrate bindings. The numerical configuration $\Theta$ and causal update rules defined in Section~\ref{sec:evaluated_monitor} remain fixed across substrates. All linguistic bindings use qwen3-embedding representations with 4,096 dimensions, served locally through Ollama. For BOIR, the estimator supplies the maximum a posteriori hypothesis for each five-step window as its attribution $i_t$. The BOIR binding therefore applies the aggregate gain gate specified in Appendix~\ref{app:stalling_index}.

\begin{table*}[!ht]
\centering
\caption{Substrate-specific AST bindings, evidence representations, and telemetry channel interpretations across all eight experimental domains. The task schema, evidence space, task-resolution map, and evidence representation vary across substrates, while the monitor configuration remains frozen.}
\label{tab:supp_bindings}

\scriptsize
\setlength{\tabcolsep}{3pt}
\renewcommand{\arraystretch}{1.15}

\begin{tabularx}{\textwidth}{@{}
  >{\RaggedRight\arraybackslash}p{0.11\textwidth}
  Y
  Y
  Y
  Y
  Y
  Y
@{}}
\toprule
\textbf{Substrate} &
$\boldsymbol{\mathcal{M}}$ &
$\boldsymbol{\mathcal{Y}}$ &
$\boldsymbol{S(y_t)}$ &
$\boldsymbol{e(y_t)}$ &
\textbf{Progress interpretation} &
\textbf{Friction interpretation} \\
\midrule

\textbf{SAR} &
Task-critical missing-person facts &
Witness/source responses to queries &
Expert completeness per fact; five-model LLM ensemble in the robustness check &
Response embedding &
Residual acquisition value across mission-critical facts &
Repeated or degraded witness evidence associated with unresolved facts \\

\addlinespace[3pt]

\textbf{AnnoMI} &
Projected client-state dimensions &
Client statements within MI sessions &
Manual rubric annotation of client disclosures &
Client utterance embedding &
Residual acquisition value across projected client-state dimensions &
Repeated low-gain client evidence associated with unresolved dimensions \\

\addlinespace[3pt]

\textbf{MultiWOZ} &
Transactional booking slots &
User utterances following system prompts &
Belief-state slot completeness changes &
User-turn text embedding &
Residual acquisition value across required booking slots &
Repeated constraints while required slots remain unresolved \\

\addlinespace[3pt]

\textbf{Map Task} &
Unique session-specific landmarks &
Giver statements following follower queries &
Manual rubric annotation of landmark grounding &
Giver utterance embedding &
Residual acquisition value across landmark coordinates &
Repeated spatial descriptions while landmark grounding remains unresolved \\

\addlinespace[3pt]

\textbf{MiniGrid} &
$3\times3$ grid-sector partition &
$K$-step spatial visitation windows &
Fraction of walkable tiles revealed &
Unit-norm visitation and action histogram &
Residual acquisition value across spatial sectors &
Recurrent local visitation associated with incomplete spatial coverage \\

\addlinespace[3pt]

\textbf{Crafter} &
22-element capability achievement tree &
Action, inventory, and achievement logs &
Fraction of unlock prerequisites satisfied &
Unit-norm action--inventory--milestone signature &
Residual acquisition value across achievement prerequisites &
Low nominal friction under predominantly atomic updates; sustained pre-resolution recurrence remains detectable \\

\addlinespace[3pt]

\textbf{MetaWorld} &
Manipulation milestones comprising reach, grasp, lift and place &
$K$-step continuous kinematic traces &
Deterministic completion from simulator states &
Unit-norm $\mathbb{R}^{13}$ hand, object, gripper, and displacement summary &
Residual acquisition value across manipulation milestones &
Recurrent kinematic evidence associated with unresolved manipulation coordinates \\

\addlinespace[3pt]

\textbf{BOIR estimator} &
BOIR goal hypotheses $h_1,\ldots,h_N$ &
Short windows of posterior dynamics and observation features &
Posterior $p_i$ of the true hypothesis and $1-p_i$ of each competitor at the end of each window &
Unit-norm posterior-dynamics signature (posterior mean and shift, entropy level and trend, mean KL step, MAP switches, heading change and reversals) &
Residual adjudication value across intent hypotheses &
Recurrent estimator evidence associated with unresolved hypothesis adjudication \\

\bottomrule
\end{tabularx}
\end{table*}

\vspace{1em}

\section{Telemetry Robustness Under Automated Scorer Substitution}
\label{supp_note_scorer}

The controlled SAR inquiry experiment in Section~\ref{sec:exp1} was repeated using automated evidence scores from five locally served language models. The ensemble comprised Gemma 3 12B, Ministral 3 14B, Gemma 2, Qwen 3 and Llama 3.1. Ollama served the models at temperature zero. Each model scored each response independently with the same task-resolution rubric. Krippendorff $\alpha$ for interval scores across turn-coordinate units was 0.827 for the hiker corpus, 0.716 for the elderly corpus and 0.853 for the child corpus. Pooled agreement was 0.801 across the five models and 0.757 after including the expert as a sixth annotator.

Automated scoring assigned lower absolute completion estimates in all three inquiry regimes (Appendix Table~\ref{tab:supp_sar_scorer}). Efficient inquiry (S1) produced mean friction of $0.04\pm0.01$ and a stall burden of $3.33\%$. Repetitive questioning (S2) produced mean friction of $0.20\pm0.05$ and a stall burden of $35.00\%$. Repetitive witness replies (S3) produced mean friction of $0.24\pm0.11$ and a stall burden of $33.33\%$. The friction ordering across regimes matched that obtained with expert scores. The monitor configuration remained fixed during scorer substitution.

\begin{table}[!ht]
\centering
\caption{AST telemetry under expert annotations and automated language-model ensemble scoring in controlled SAR inquiry. S1 denotes efficient sequential acquisition. S2 denotes repetitive agent questioning. S3 denotes repetitive witness responses. Values are means and standard deviations across three mission scenarios.}
\label{tab:supp_sar_scorer}

\footnotesize
\setlength{\tabcolsep}{6pt}
\renewcommand{\arraystretch}{1.15}

\begin{tabular}{@{}lcccccc@{}}
\toprule
\multirow{2}{*}{\textbf{Regime}} &
\multicolumn{3}{c}{\textbf{Human Expert Scorer}} &
\multicolumn{3}{c}{\textbf{Automated LLM Ensemble Scorer}} \\
\cmidrule(lr){2-4} \cmidrule(lr){5-7}
& \makecell{\textbf{Stall}\\\textbf{burden (\%)}} & \makecell{\textbf{Mean}\\$\boldsymbol{\mathrm{SI}^{\perp}}$} & \makecell{\textbf{Categories}\\\textbf{resolved}}
& \makecell{\textbf{Stall}\\\textbf{burden (\%)}} & \makecell{\textbf{Mean}\\$\boldsymbol{\mathrm{SI}^{\perp}}$} & \makecell{\textbf{Categories}\\\textbf{resolved}} \\
\midrule
S1 & $\mathbf{0.00 \pm 0.00}$ & $\mathbf{0.02 \pm 0.004}$ & $\mathbf{8.00 \pm 0.00}$ & $\mathbf{3.33 \pm 4.71}$ & $\mathbf{0.04 \pm 0.01}$ & $\mathbf{4.67 \pm 1.25}$ \\
S2 & $35.00 \pm 7.07$ & $0.26 \pm 0.08$ & $4.33 \pm 0.94$ & $35.00 \pm 4.08$ & $0.20 \pm 0.05$ & $2.00 \pm 0.82$ \\
S3 & $36.67 \pm 12.47$ & $0.28 \pm 0.09$ & $4.67 \pm 1.25$ & $33.33 \pm 14.34$ & $0.24 \pm 0.11$ & $3.00 \pm 0.82$ \\
\bottomrule
\end{tabular}
\end{table}

\vspace{1em}

\section{Additional Behavioral Analyses in Human Interaction}
\label{supp_note_human}

The AnnoMI evaluation in Section~\ref{sec:exp4} maps client statements from 20 motivational interviewing sessions onto five declared client-state dimensions. These dimensions represent the presenting problem, motivation, barriers, resources and action readiness. Appendix Table~\ref{tab:supp_annomi} reports Spearman correlations between friction and acquisition throughput. Greater friction was associated with lower information gain per turn and lower cumulative task resolution within these declared dimensions.

\begin{table}[!ht]
\centering
\caption{Spearman rank correlations between AST friction measures and information acquisition in the 20 AnnoMI motivational interviewing sessions.}
\label{tab:supp_annomi}

\footnotesize
\setlength{\tabcolsep}{8pt}
\renewcommand{\arraystretch}{1.15}

\begin{tabular}{@{}lcc@{}}
\toprule
\textbf{Diagnostic metric pair} & \textbf{Spearman $\boldsymbol{\rho}$} & \textbf{$p$-value} \\
\midrule
Stall burden vs. information gain / turn & $-0.666$ & $0.0014$ \\
Mean friction ($\mathrm{SI}^{\perp}$) vs. information gain / turn & $-0.666$ & $0.0013$ \\
Stall burden vs. total knowledge resolved & $-0.570$ & $0.0087$ \\
Mean friction ($\mathrm{SI}^{\perp}$) vs. total knowledge resolved & $-0.646$ & $0.0021$ \\
\bottomrule
\end{tabular}
\end{table}

\vspace{1em}

\section{Extended Evaluation on Non-Linguistic and Estimator Substrates}
\label{supp_note_nonling}

\subsection{Progress prediction across forecast horizons}
\label{app:progress_horizons}
Appendix Table~\ref{tab:supp_progress_acc} reports top-1 next-gain prediction accuracy for the Bayesian Progress Estimator $\mathbf{PE}^{\mathrm{B}}$ at horizons $k\in\{1,3,5\}$. The comparison methods are uniform guessing and recency-based prediction.

On clean Crafter and MetaWorld trajectories, the progress channel achieved top-1 accuracies of $0.544$ and $0.985$, respectively, at $k=5$. Injected loops reduced accuracy in both substrates, with horizon-specific changes in the strongest baseline. In MiniGrid, recency-based prediction exceeded $\mathbf{PE}^{\mathrm{B}}$ at all three horizons, consistent with locally coupled spatial transitions.

\begin{table*}[!ht]
\centering
\caption{Top-1 next-gain prediction accuracy of the Bayesian Progress Estimator $\mathrm{PE}^{\mathrm{B}}$ and two baselines on non-linguistic substrates at horizons $k\in\{1,3,5\}$. Bold values identify the highest accuracy within each substrate and horizon.}
\label{tab:supp_progress_acc}

\footnotesize
\setlength{\tabcolsep}{6pt}
\renewcommand{\arraystretch}{1.15}

\begin{tabular*}{\textwidth}{@{\extracolsep{\fill}}lccccccccc@{}}
\toprule
\multirow{2}{*}{\textbf{Substrate}} &
\multicolumn{3}{c}{\textbf{PE$^{\mathrm{B}}$ (Ours)}} &
\multicolumn{3}{c}{\textbf{Recency Baseline}} &
\multicolumn{3}{c}{\textbf{Uniform Guessing}} \\
\cmidrule(lr){2-4} \cmidrule(lr){5-7} \cmidrule(l){8-10}
& \textbf{$k{=}1$} & \textbf{$k{=}3$} & \textbf{$k{=}5$}
& \textbf{$k{=}1$} & \textbf{$k{=}3$} & \textbf{$k{=}5$}
& \textbf{$k{=}1$} & \textbf{$k{=}3$} & \textbf{$k{=}5$} \\
\midrule
MiniGrid & 0.253 & 0.326 & 0.373 & \textbf{0.603} & \textbf{0.624} & \textbf{0.651} & 0.114 & 0.155 & 0.235 \\
Crafter (clean) & \textbf{0.226} & \textbf{0.380} & \textbf{0.544} & 0.077 & 0.178 & 0.228 & 0.090 & 0.129 & 0.211 \\
Crafter (with loops) & \textbf{0.132} & 0.119 & \textbf{0.100} & 0.118 & \textbf{0.147} & 0.086 & 0.118 & 0.046 & 0.071 \\
MetaWorld (clean) & \textbf{0.906} & \textbf{0.977} & \textbf{0.985} & 0.531 & 0.360 & 0.333 & 0.333 & 0.320 & 0.294 \\
MetaWorld (with loops) & 0.643 & \textbf{0.819} & \textbf{0.893} & \textbf{0.685} & 0.573 & 0.593 & 0.274 & 0.313 & 0.390 \\
\bottomrule
\end{tabular*}
\end{table*}

\subsection{Internal monitoring of the BOIR intent estimator}
\label{app:boir_internal}
The BOIR evaluation in Section~\ref{sec:boir_estimator} binds AST to an internal Bayesian estimator of human goals. Appendix Table~\ref{tab:supp_boir} reports telemetry from 100 evaluation epochs, comprising 40 epochs from 20 two-epoch S1 trials and 60 epochs from 20 three-epoch S3 trials. Friction rose within epochs during hypothesis re-adjudication and remained low immediately after goal changes. Each goal change resets the acquisition state and evidence history.

\begin{table}[!ht]
\centering
\caption{AST telemetry in the BOIR Bayesian intent estimator. Confident incorrect starts are epochs with an initially incorrect maximum a posteriori hypothesis whose posterior probability exceeds $0.70$. True-hypothesis resolution reports the mean posterior assigned to the correct human goal at the beginning and end of each epoch.}
\label{tab:supp_boir}

\footnotesize
\setlength{\tabcolsep}{6pt}
\renewcommand{\arraystretch}{1.15}

\begin{tabular}{@{}lccccc@{}}
\toprule
\textbf{Scenario} &
\makecell{\textbf{Initial MAP} \\ \textbf{errors}} &
\makecell{\textbf{Confident} \\ \textbf{wrong starts}} &
\makecell{\textbf{True-hypothesis} \\ \textbf{resolution}} &
\makecell{$\boldsymbol{\mathrm{SI}^{\perp}}$ \\ \textbf{post-change}} &
\makecell{$\boldsymbol{\mathrm{SI}^{\perp}}$ \\ \textbf{mid-epoch}} \\
\midrule
S1 & 20/40 & 4/40 & $0.467 \rightarrow 0.834$ & 0.014 & 0.051 \\
S3 & 60/60 & 42/60 & $0.142 \rightarrow 0.966$ & 0.075 & 0.130 \\
\bottomrule
\end{tabular}
\end{table}

\vspace{1em}

\section{Closed-Loop Reinforcement Learning Hyperparameters and Ablations}
\label{supp_note_rl}

The closed-loop experiment in Section~\ref{sec:exp5} trains five PPO policy variants with 10 random seeds for each of the three SAR corpora and two termination conditions. The corpus consists of child, elderly and hiker scenarios. This design yields 300 trained policies.

\subsection{Environment and observations}
\label{app:rl_environment}
Each episode contains at most 50 interview exchanges. An action selects a questioning strategy and a target requirement or selects a general question. The environment returns the corpus response, expert resolution scores and evidence embedding. The frozen monitor then updates its causal state. Each exchange incurs a cost of 0.1. An episode ends with a bonus of 10 once every coordinate satisfies $\upsilon_i\geq\upsilon_{\mathrm{complete}}$. All policy observations include $\boldsymbol{\upsilon}(t)$ and the elapsed fraction of the interaction budget. AST-Pen additionally receives $\mathbf{PE}^{\mathrm{B}}(t)$. AST-Obs and Full-AST receive both $\mathbf{PE}^{\mathrm{B}}(t)$ and $\mathrm{SI}^{\perp}(t)$. AST-Pen and Full-AST receive an exchange-level reward penalty equal to $\mathrm{SI}^{\perp}(t)$. Condition B terminates the episode at the first exchange satisfying $\mathrm{SI}^{\perp}(t)>\theta_{\mathrm{SI}}$.

\subsection{Operational-cost comparator}
\label{app:rl_operational_cost}
The Coupled-Blind comparator receives no telemetry and incurs an operational cost $\rho c(t)$. The cost $c(t)$ adds two units for repeating an action already taken during the episode and two units when aggregate acquired resolution is below $10^{-3}$. A uniform random reference policy was evaluated over 200 Condition A episodes per corpus to calibrate $\rho$ before training. Mean friction was 0.314 and mean operational cost was 2.19 per exchange, yielding $\rho=0.143$. This calibration matches the mean penalty magnitude under the reference policy.

\subsection{Policy training and evaluation}
\label{app:rl_training}
All agents use a Stable-Baselines3 actor-critic multilayer perceptron with two hidden layers of 64 units and tanh activations. Policies are trained for 50,000 environment steps with learning rate $3\times10^{-4}$, rollout length 256, minibatch size 64, 10 epochs per update, discount factor $\gamma=0.99$, generalized advantage estimation parameter $\lambda_{\mathrm{GAE}}=0.95$, clip ratio $0.2$ and entropy coefficient $0.02$. Training uses fixed seeds and one CPU thread. Greedy policy evaluation in the deterministic environment uses one episode per trained policy. Reported confidence intervals are bootstrapped over seed means using 10,000 resamples.

\vspace{1em}

\section{Architectural Notation and Contract Glossary}
\label{supp_note_notation}

Appendix Table~\ref{tab:supp_notation} defines the symbols and data types used in the task binding, acquisition state and frozen monitor described in Section~\ref{sec:methods_detailed}.

\begin{longtable}{@{}
  >{\RaggedRight\arraybackslash}p{0.16\textwidth}
  >{\RaggedRight\arraybackslash}p{0.22\textwidth}
  >{\RaggedRight\arraybackslash}p{0.56\textwidth}
@{}}
\caption{Core notation and interface definitions for Acquisition-State Telemetry.}
\label{tab:supp_notation}\\
\toprule
\textbf{Symbol} & \textbf{Type / Domain} & \textbf{Operational Definition} \\
\midrule
\endfirsthead
\multicolumn{3}{l}{\tablename~\thetable\ continued}\\
\toprule
\textbf{Symbol} & \textbf{Type / Domain} & \textbf{Operational Definition} \\
\midrule
\endhead
\midrule
\multicolumn{3}{r}{Continued on next page}\\
\endfoot
\bottomrule
\endlastfoot

$\mathcal{B}$ & Interface Tuple & Task binding $\mathcal{B} = (\mathcal{M}, \mathcal{Y}, S, e)$. \\
$\mathcal{M}$ & Finite Set & Schema coordinates $\{1, \ldots, N\}$ representing task dimensions. \\
$\mathcal{Y}$ & Observation Space & Evidence space emitted by the monitored process. \\
$y_t$ & Element $\in \mathcal{Y}$ & Evidence observation received at exchange $t$. \\
$i_t$ & $\mathcal{M} \cup \{\bot\}$ & Requirement targeted by the preceding action. The marker $\bot$ denotes an unattributed exchange. \\
$S$ & Function $\mathcal{Y} \to [0,1]^N$ & Task-resolution map projecting evidence onto schema resolution. \\
$\mathbf{s}(t)$ & Vector $\in [0,1]^N$ & Instantaneous resolution observation $S(y_t) = (s_1(t), \ldots, s_N(t))$. \\
$e$ & Function $\mathcal{Y} \to \mathbb{R}^d$ & Evidence representation map with unit-norm output ($\lVert e(y)\rVert_2 = 1$). \\
$\mathbf{r}^{(t)}$ & Vector $\in \mathbb{R}^d$ & Unit-norm embedding of current evidence for subspace comparison. \\
$\upsilon_i(t)$ & Scalar $\in [0, 1]$ & Cumulative attained resolution of coordinate $i$ ($\max_{\ell \le t} s_i(\ell)$). \\
$\Delta\upsilon_i(t)$ & Scalar $\ge 0$ & Marginal resolution gain acquired at exchange $t$. \\
$\tau_{1:t}$ & Ordered Sequence & Causal acquisition history available through exchange $t$. \\
$\Phi_\Theta$ & Functional & Frozen causal monitor mapping history $\tau_{1:t}$ to telemetry $z_t$. \\
$\Theta$ & Parameter Tuple & Frozen internal monitor configuration and numerical thresholds. \\
$\mathbf{P}(t)$ & Vector $\in \mathbb{R}_{\ge 0}^N$ & Prospective acquisition value for each declared requirement. \\
$F(t)$ & Scalar $\in [0, 1]$ & Friction channel reporting gain-gated representational recurrence. \\
$z_t$ & Tuple $(\mathbf{P}(t), F(t))$ & Live telemetry tuple emitted at exchange $t$. \\
$\mathbf{PE}^{\mathrm{B}}(t)$ & Estimator Vector & Reference Bayesian Progress Estimator instantiation. \\
$\mathrm{SI}^\perp(t)$ & Index $\in [0, 1]$ & Reference gain-gated Stalling Index via orthogonal projection. \\
$\pi$ & Policy Function & Downstream controller conditioning actions on $z_t$. \\
\end{longtable}

\clearpage

\section{Formal Task-Referenced Acquisition Model}
\label{supp_note_formal}

\subsection{Task binding and marked evidence}
\label{app:formal_binding}
Let $\mathcal{M}=\{1,\ldots,N\}$ denote the finite set of task requirements and $\mathcal{Y}$ the observation space. Let $S:\mathcal{Y}\to[0,1]^N$ map observations to task resolution and $e:\mathcal{Y}\to\mathbb{R}^{d}$ map observations to unit-norm comparative representations. The task binding is $\mathcal{B}=(\mathcal{M},\mathcal{Y},S,e)$. The marked observation $\chi_t=(y_t,i_t)$ pairs the incoming evidence with the requirement targeted by the preceding action. The attribution marker satisfies $i_t\in\mathcal{M}\cup\{\bot\}$, where $\bot$ denotes an unattributed exchange. The two projections are
\begin{equation}
\mathbf{s}(t)=S(y_t)\in[0,1]^N,\qquad
\mathbf r^{(t)}=e(y_t),\quad \lVert\mathbf r^{(t)}\rVert_2=1.
\label{eq:formal_dual_projection}
\end{equation}
Attribution identifies the target of an action. Its evidence may resolve any declared requirement, including requirements outside the target and requirements acquired during unattributed exchanges.

\subsection{Acquisition state and admissibility}
\label{app:formal_state}
The attained-resolution state accumulates monotonically within a monitoring episode according to
\begin{equation}
\upsilon_i(t)=\max\{\upsilon_i(t-1),s_i(t)\},\qquad \upsilon_i(0)=0,
\quad \Delta\upsilon_i(t)=\upsilon_i(t)-\upsilon_i(t-1).
\label{eq:formal_state_accumulation}
\end{equation}
The signed increment $\delta_i(t)=s_i(t)-\upsilon_i(t-1)$ records the difference between the current evidence score and the previously attained resolution. Each requirement retains a stable interpretation during the episode. Invalidation of verified evidence terminates that episode and resets the attained state and evidence histories. The causal monitor processes an ordered acquisition trace defined by
\begin{equation}
\tau_{1:t}=\bigl((\chi_\ell,\mathbf s(\ell),\boldsymbol\upsilon(\ell),\Delta\boldsymbol\upsilon(\ell))\bigr)_{\ell=1}^{t},\qquad
z_t=\Phi_{\Theta}(\tau_{1:t})=(\mathbf P(t),F(t)).
\label{eq:formal_monitor_contract}
\end{equation}
The monitor emits prospective acquisition values $\mathbf P(t)\in\mathbb R_{\ge0}^{N}$ and scalar friction $F(t)\in[0,1]$. Appendix~\ref{supp_note_monitor_math} specifies the frozen reference implementation used in the experiments.

\subsection{Gain-gated recurrence}
\label{app:formal_recurrence}
The friction channel combines recurrence against historical attributed representations, a realized-gain gate and the remaining unresolved fraction of each requirement. The evaluated reference implementation follows the structural expression
\begin{equation}
F(t)\simeq \sum_{i\in\mathcal A_t} \omega_i(t)\,
\underbrace{\mu_i(t)}_{\text{recurrence}}\,
\underbrace{D(\varrho(t))}_{\text{gain gate}}\,
\underbrace{(1-\upsilon_i(t))}_{\text{unresolved volume}},
\qquad \omega_i(t)=\frac{\mu_i(t)}{\sum_{j\in\mathcal A_t}\mu_j(t)},
\label{eq:formal_friction_structure}
\end{equation}
The implemented index includes numerical regularization, clipping and a zero-history convention, as specified in Eq.~\ref{eq:si_complete_methods}. Positive realized gain attenuates recurrent evidence. Completed requirements contribute zero through their unresolved-volume factor. Recurrent evidence can raise friction when targeted gain is negligible or negative and the associated requirement remains incomplete.

\section{Full Computational Specification of the Frozen Monitor}
\label{supp_note_monitor_math}

The frozen reference monitor implements the acquisition contract specified in Appendix~\ref{supp_note_formal}. Its causal update rules and numerical configuration are shared across all eight substrate bindings. The equations below define the evaluated estimator and friction index.

\subsection{Signed increments, counters and histories}
\label{app:monitor_histories}
At exchange $t$, for each coordinate $i\in\mathcal{M}$, the signed increment relative to the attained state and the realized gain are
\begin{equation}
\delta_i(t) = s_i(t) - \upsilon_i(t-1), \qquad \Delta\upsilon_i(t) = \max\bigl(0,\, \delta_i(t)\bigr).
\label{eq:signed_increment_methods}
\end{equation}
The monitor updates attributed attempt counts $m_i(t)$ and material success counts $k_i(t)$ according to
\begin{align}
m_i(t) &= m_i(t-1) + \mathbb{I}[i_t = i], \nonumber\\
k_i(t) &= k_i(t-1) + \mathbb{I}[i_t = i]\,\mathbb{I}[\delta_i(t) > \tau_{\mathrm{g}}],
\label{eq:counts_methods}
\end{align}
where $\tau_{\mathrm{g}}$ is the material-success threshold. Each coordinate keeps a gain-weighted evidence path
\begin{equation}
\mathbf{c}_i(t) = \mathbf{c}_i(t-1) + \Delta\upsilon_i(t)\,\mathbf{r}^{(t)}, \qquad \mathbf{c}_i(0)=\mathbf{0},
\label{eq:track_methods}
\end{equation}
and records its value after every positive-gain event. With $G_i(t)$ the number of such events through exchange $t$, the checkpoint sequence is $\mathcal{C}_i(t) = (\mathbf{c}_i^{(0)}, \ldots, \mathbf{c}_i^{(G_i(t))})$, with $\mathbf{c}_i^{(0)}=\mathbf{0}$. For friction estimation, each requirement retains a history of representations attributed to it, given by
\begin{equation}
\mathcal{R}_i(t) = \bigl\{\mathbf{r}^{(\ell)} : 1 \leq \ell \leq t,\; i_\ell = i\bigr\}.
\label{eq:recurrence_history_methods}
\end{equation}

\subsection{Bayesian Progress Estimator}
\label{app:progress_estimator}
The yield $\theta_i$ of an attributed attempt on coordinate $i$ has a uniform $\operatorname{Beta}(1,1)$ prior, so that
\begin{equation}
\theta_i(t) \mid m_i(t), k_i(t) \sim \operatorname{Beta}\bigl(1 + k_i(t),\, 1 + m_i(t) - k_i(t)\bigr),
\label{eq:beta_methods}
\end{equation}
with posterior mean and variance
\begin{equation}
\begin{aligned}
\mathbb{E}[\theta_i(t)]
&= \frac{1 + k_i(t)}{2 + m_i(t)}, \\[6pt]
\sigma_i^2(t)
&= \frac{(1 + k_i(t))(1 + m_i(t) - k_i(t))}
{(2 + m_i(t))^2 (3 + m_i(t))}.
\end{aligned}
\label{eq:moments_methods}
\end{equation}
The uncertainty multiplier gives greater weight to coordinates with more uncertain estimated attempt yields and is defined by
\begin{equation}
\epsilon_i(t) = 1 + \beta\, \frac{\sigma_i^2(t)}{\max\left(\varepsilon_{\mathrm{P}},\, \max_{j\in\mathcal{M}} \sigma_j^2(t)\right)} \in [1,\, 1+\beta].
\label{eq:epistemic_methods}
\end{equation}
Let $n_i(t)=\min(n,G_i(t)+1)$ and $G=G_i(t)$. The straightness of the most recent checkpoints is the ratio of net displacement to traversed path length, defined by
\begin{equation}
\varsigma_i(t) = \frac{\left\|\mathbf{c}_i^{(G)} - \mathbf{c}_i^{(G - n_i(t) + 1)}\right\|_2}{\sum_{j=G-n_i(t)+2}^{G} \left\|\mathbf{c}_i^{(j)} - \mathbf{c}_i^{(j-1)}\right\|_2 + \varepsilon_{\mathrm{G}}} \in [0, 1].
\label{eq:ratio_methods}
\end{equation}
The trajectory term is $\xi_i(t) = 1 - \min\bigl(1, \upsilon_i(t)/\upsilon^\star\bigr)\,\varsigma_i(t)$, with $\xi_i(t)=1$ before the first gain event ($G_i(t)=0$). Declared prerequisites gate progress through $g_i(t) = \prod_{j \in \operatorname{dep}(i)} \mathbb{E}[\theta_j(t)]$, with $g_i(t)=1$ when $\operatorname{dep}(i)=\varnothing$. With binary entropy $H_b(\upsilon) = -\upsilon\log_2\upsilon - (1-\upsilon)\log_2(1-\upsilon)$, evaluated on $\upsilon_i(t)$ clipped to $[\varepsilon_{\mathrm{P}}, 1-\varepsilon_{\mathrm{P}}]$, the progress channel is
\begin{equation}
\begin{aligned}
\mathrm{PE}^{\mathrm{B}}_i(t)
= w_i\, g_i(t) \Bigl[\,
&\alpha\,\mathbb{E}[\theta_i(t)]\,H_b\bigl(\upsilon_i(t)\bigr)\,\epsilon_i(t) \\
&+ (1-\alpha)\,\xi_i(t)
\Bigr].
\end{aligned}
\label{eq:pe_complete_methods}
\end{equation}
The first term combines the posterior mean attempt yield, binary entropy of attained resolution and yield uncertainty. Its contribution approaches a numerical floor near full resolution because the entropy argument is clipped for stability. The second term decreases as the gain-weighted evidence path becomes straighter and attained resolution approaches its trajectory-normalization scale $\upsilon^\star$. Circuitous paths and low attained resolution retain larger trajectory contributions.

\subsection{Gain-gated Stalling Index}
\label{app:stalling_index}
The representations of earlier exchanges attributed to coordinate $i$, $\mathcal{R}_i(t-1)$, form the columns of the history matrix $\mathbf{V}_i(t^-)$. The projector onto the orthogonal complement of the historical subspace is
\begin{equation}
\boldsymbol{\Pi}_i^\perp(t) = \mathbf{I} - \mathbf{V}_i(t^-)\mathbf{V}_i(t^-)^\dagger,
\label{eq:projector_methods}
\end{equation}
The operator $(\cdot)^\dagger$ denotes the Moore-Penrose pseudoinverse. Orthogonal novelty and its complementary recurrence are defined by
\begin{equation}
\eta_i(t) = \operatorname{clip}_{[0,1]}\left(\left\|\boldsymbol{\Pi}_i^\perp(t)\mathbf{r}^{(t)}\right\|_2\right), \qquad \mu_i(t) = 1 - \eta_i(t).
\label{eq:novelty_methods}
\end{equation}
Realized gain gates recurrence through
\begin{equation}
\begin{gathered}
D\bigl(\varrho(t)\bigr) = \max\bigl(D_{\min},\, 1 - \lambda\varrho(t)\bigr),\\
\varrho(t) =
\begin{cases}
\delta_{i_t}(t), & i_t \in \mathcal{M},\\
\sum_{i} \Delta\upsilon_i(t), & i_t = \bot.
\end{cases}
\end{gathered}
\label{eq:dampening_methods}
\end{equation}
For directly attributed exchanges, negative signed gain increases the gating factor above one and amplifies recurrence. The factor $D$ lies in $[D_{\min},1+\lambda]$. Bindings with inferred attribution use aggregate realized gain $\varrho(t)=\sum_i\Delta\upsilon_i(t)$ at every exchange. BOIR uses this convention because its attribution $i_t$ is the maximum a posteriori hypothesis. Let $\mathcal{A}_t=\{i\in\mathcal{M}\mid\mathcal{R}_i(t-1)\neq\varnothing\}$ denote the active requirements. The implemented friction index is
\begin{equation}
\mathrm{SI}^{\perp}(t)
= \operatorname{clip}_{[0,1]}\!\left[
\frac{\sum_{i\in\mathcal{A}_t} \mu_i(t)^2\, D\bigl(\varrho(t)\bigr)\bigl(1-\upsilon_i(t)\bigr)}
{\sum_{i\in\mathcal{A}_t} \mu_i(t) + \varepsilon_{\mathrm{G}}}
\right],
\label{eq:si_complete_methods}
\end{equation}
The index is set to $\mathrm{SI}^{\perp}(t)=0$ when $\mathcal{A}_t=\varnothing$ or $\sum_{i\in\mathcal{A}_t}\mu_i(t)<\varepsilon_{\mathrm{G}}$. Each active requirement contributes recurrence weighted by its relative recurrence, gain gate and remaining unresolved fraction. Numerical regularization stabilizes the denominator and clipping bounds the resulting index in $[0,1]$. Friction therefore increases with recurrent evidence that yields little resolution while task requirements remain open.

\clearpage
\section{Example Telemetry Trajectories Across Substrates}
\label{supp_note_trajectories}

\begin{figure}[!p]
\centering
\includegraphics[width=\textwidth,height=0.8\textheight,keepaspectratio]{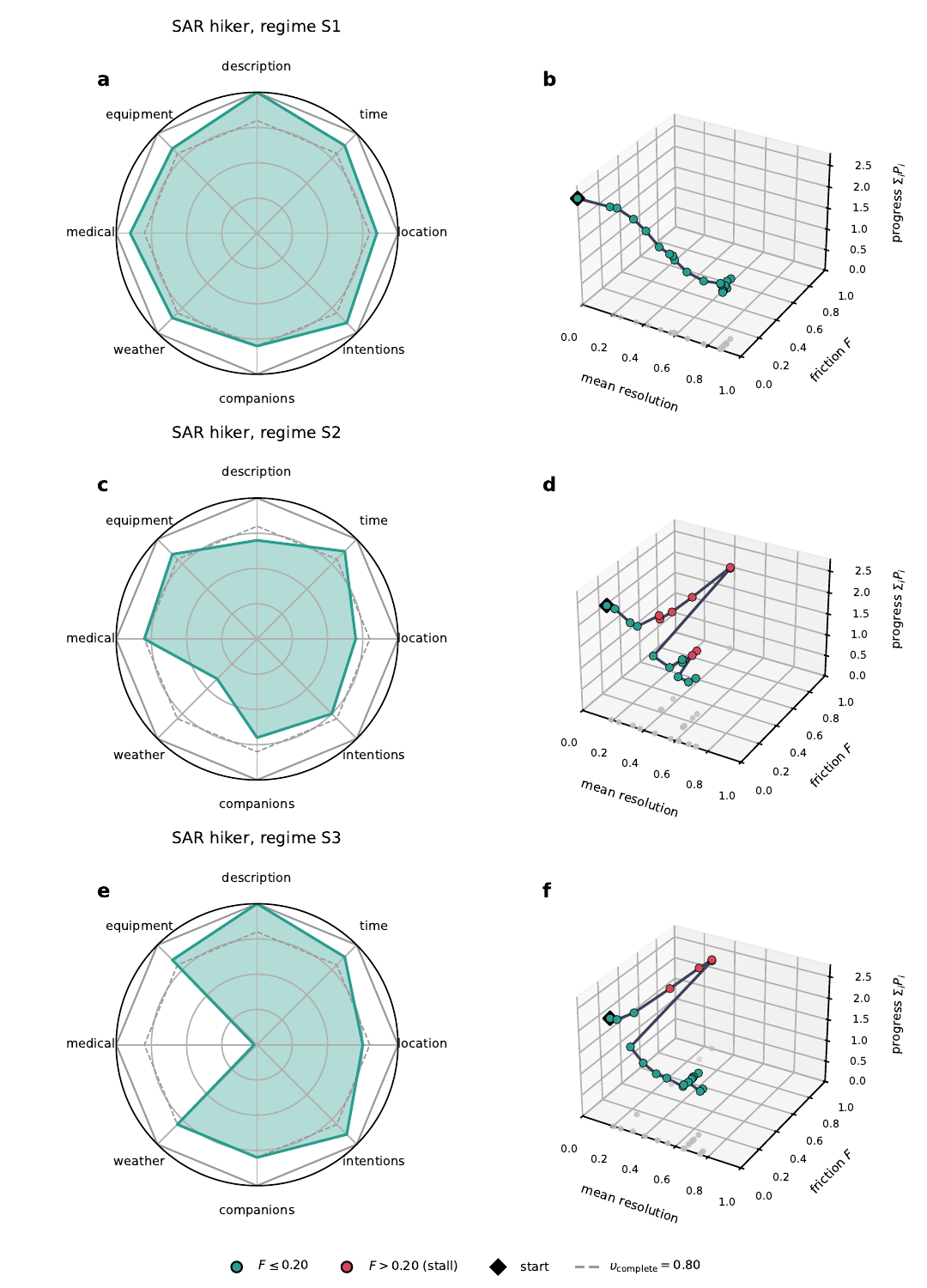}
\caption{\textbf{SAR hiker corpus with expert scoring.} Efficient inquiry (S1) resolves all eight requirements without a stalled exchange. Repetitive questioning (S2) has a stall burden of $40\%$ and resolves three requirements. Repetitive witness responses (S3) have a stall burden of $20\%$ and resolve six requirements.}
\label{fig:supp_sar_hiker}
\end{figure}

\begin{figure}[!p]
\centering
\includegraphics[width=\textwidth,height=0.8\textheight,keepaspectratio]{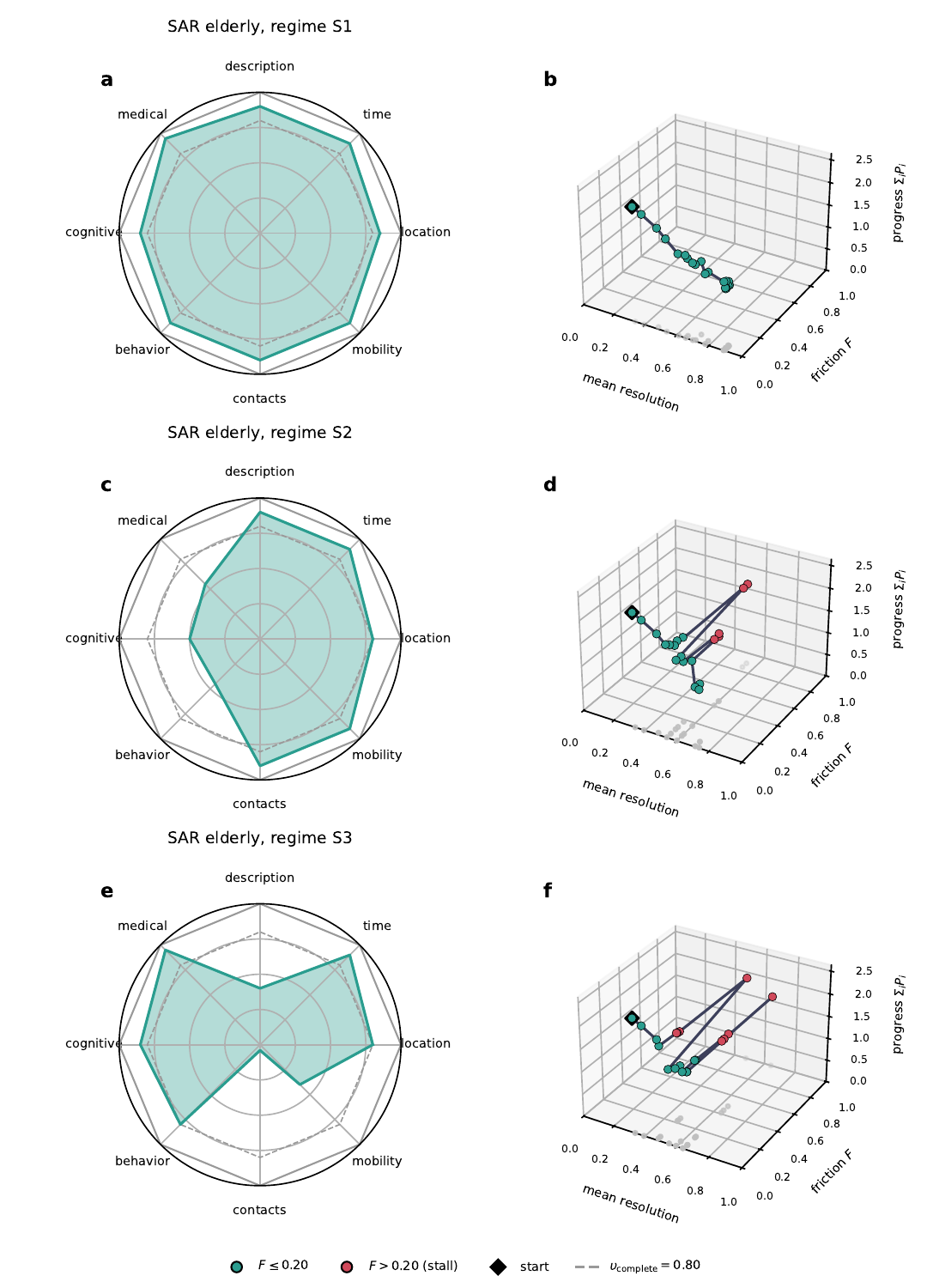}
\caption{\textbf{SAR elderly corpus with expert scoring.} Efficient inquiry (S1) resolves all eight requirements without a stalled exchange. Repetitive questioning (S2) and repetitive witness responses (S3) resolve five requirements each and have stall burdens of $25\%$ and $40\%$, respectively.}
\label{fig:supp_sar_elderly}
\end{figure}

\begin{figure}[!p]
\centering
\includegraphics[width=\textwidth,height=0.8\textheight,keepaspectratio]{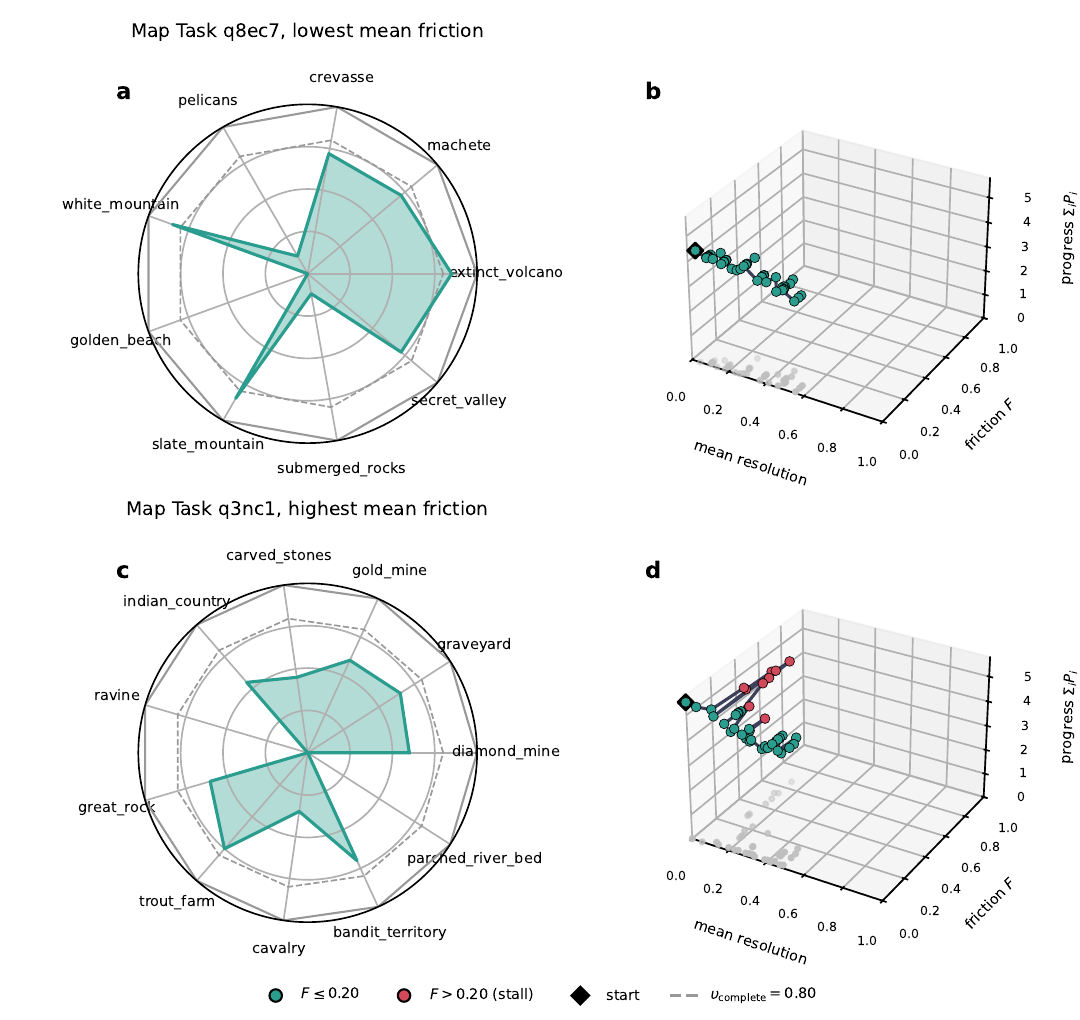}
\caption{\textbf{HCRC Map Task.} Session q8ec7 has the lowest mean friction among 27 evaluated dialogues, with $0.046$ and no stalled exchange. Session q3nc1 has the highest mean friction, with $0.137$ and a stall burden of $22\%$.}
\label{fig:supp_maptask}
\end{figure}

\begin{figure}[!p]
\centering
\includegraphics[width=\textwidth,height=0.8\textheight,keepaspectratio]{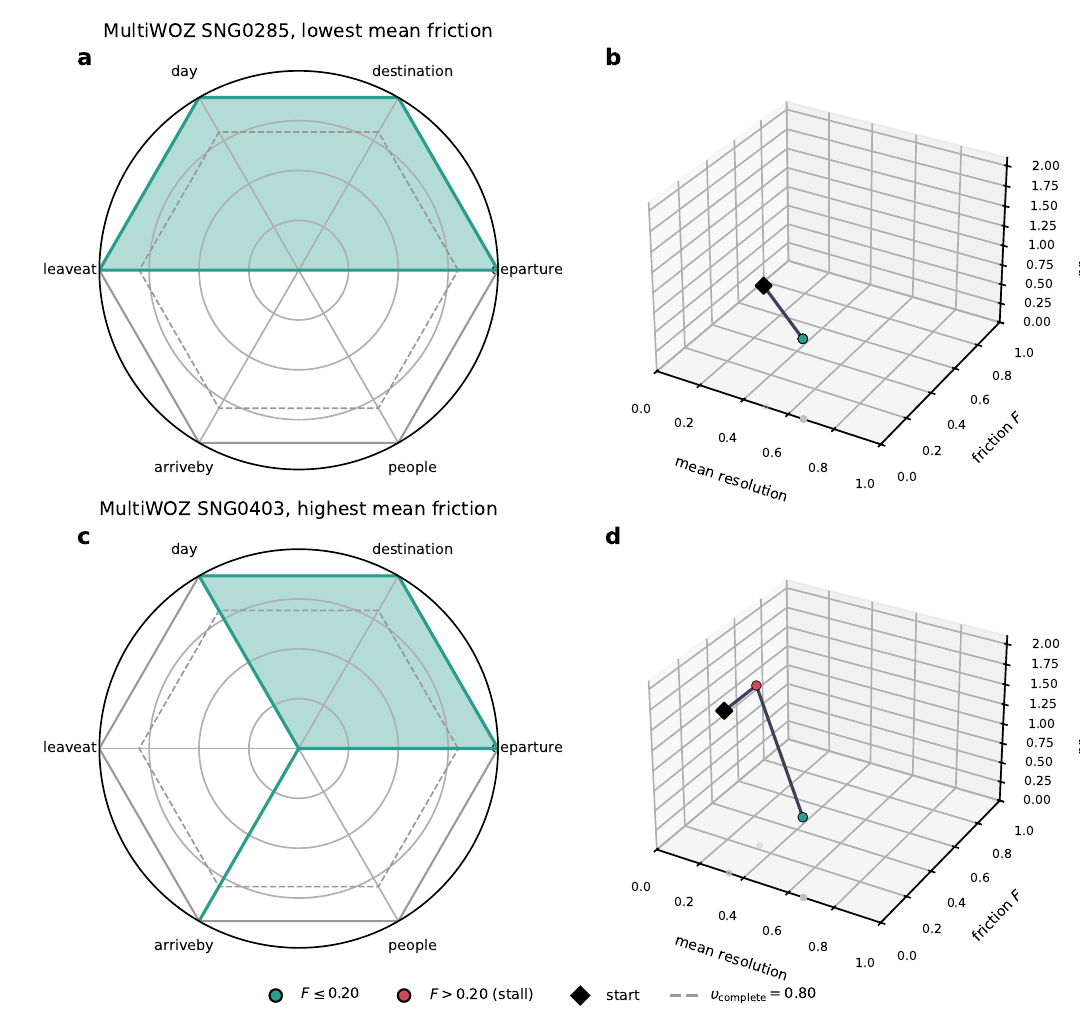}
\caption{\textbf{MultiWOZ.} Session SNG0285 contains no stalled exchange. Session SNG0403 is the only evaluated session with a stalled exchange, occurring once in four turns. Required slots resolve in single-turn updates.}
\label{fig:supp_multiwoz}
\end{figure}

\begin{figure}[!p]
\centering
\includegraphics[width=\textwidth,height=0.8\textheight,keepaspectratio]{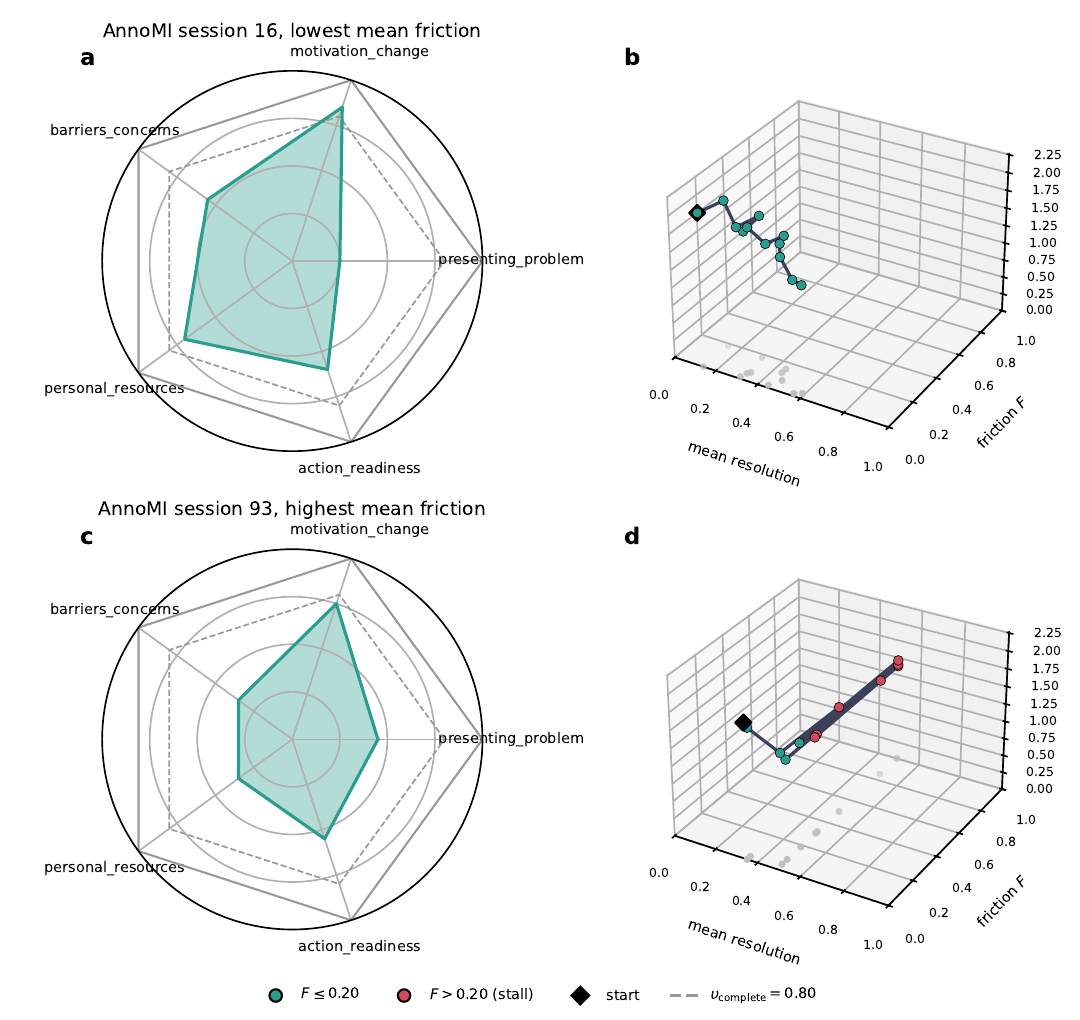}
\caption{\textbf{AnnoMI.} Session 16 has the lowest mean friction among 20 evaluated sessions, with $0.075$ and no stalled exchange. Session 93 has the highest mean friction, with $0.442$ and a stall burden of $58\%$. Both sessions received high clinical-quality ratings.}
\label{fig:supp_annomi}
\end{figure}

\begin{figure}[!p]
\centering
\includegraphics[width=\textwidth,height=0.8\textheight,keepaspectratio]{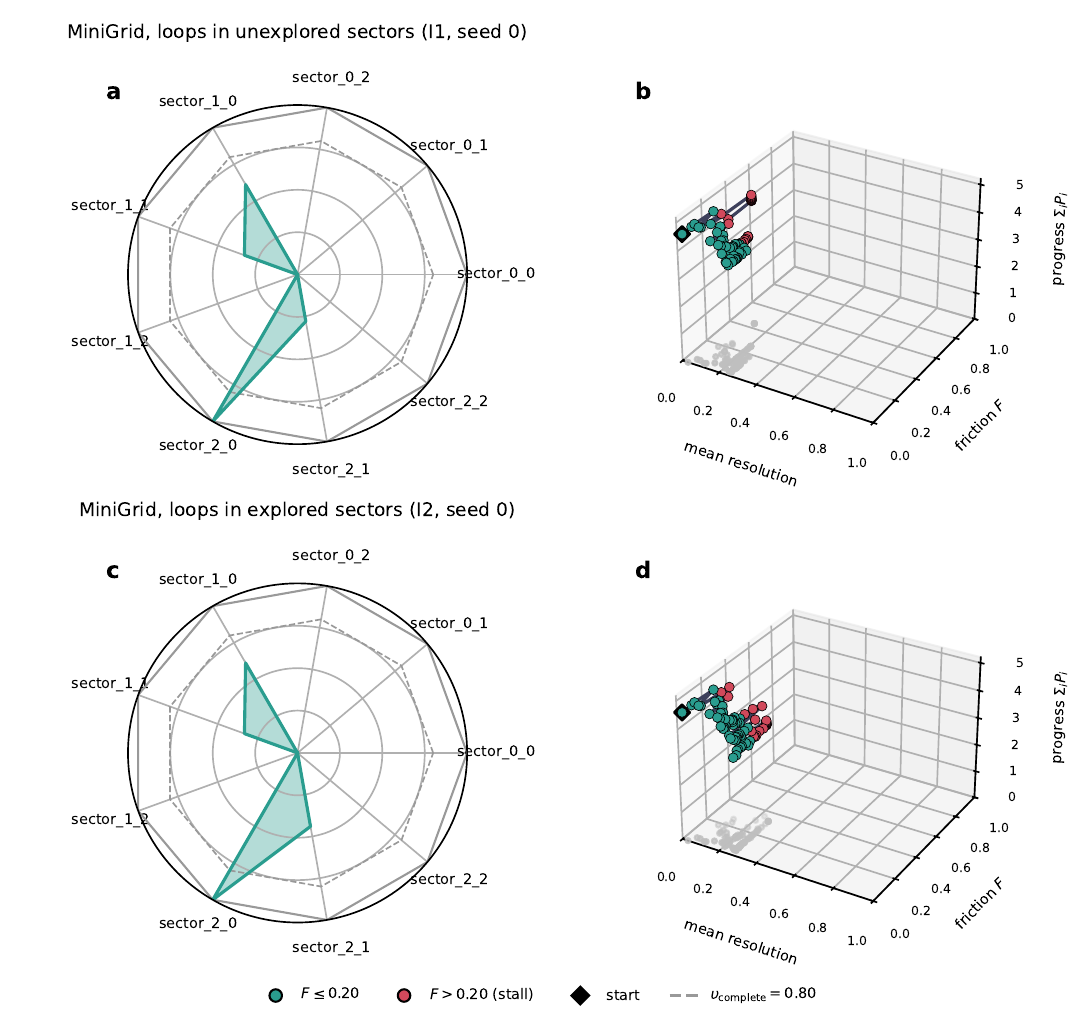}
\caption{\textbf{MiniGrid, seed 0.} Panels \textbf{a} and \textbf{b} show Condition I1, with loops in incomplete sectors during exchanges 8 through 19 and 40 through 49. Panels \textbf{c} and \textbf{d} show Condition I2, with loops in completed sectors during exchanges 55 through 66.}
\label{fig:supp_minigrid}
\end{figure}

\begin{figure}[!p]
\centering
\includegraphics[width=\textwidth,height=0.8\textheight,keepaspectratio]{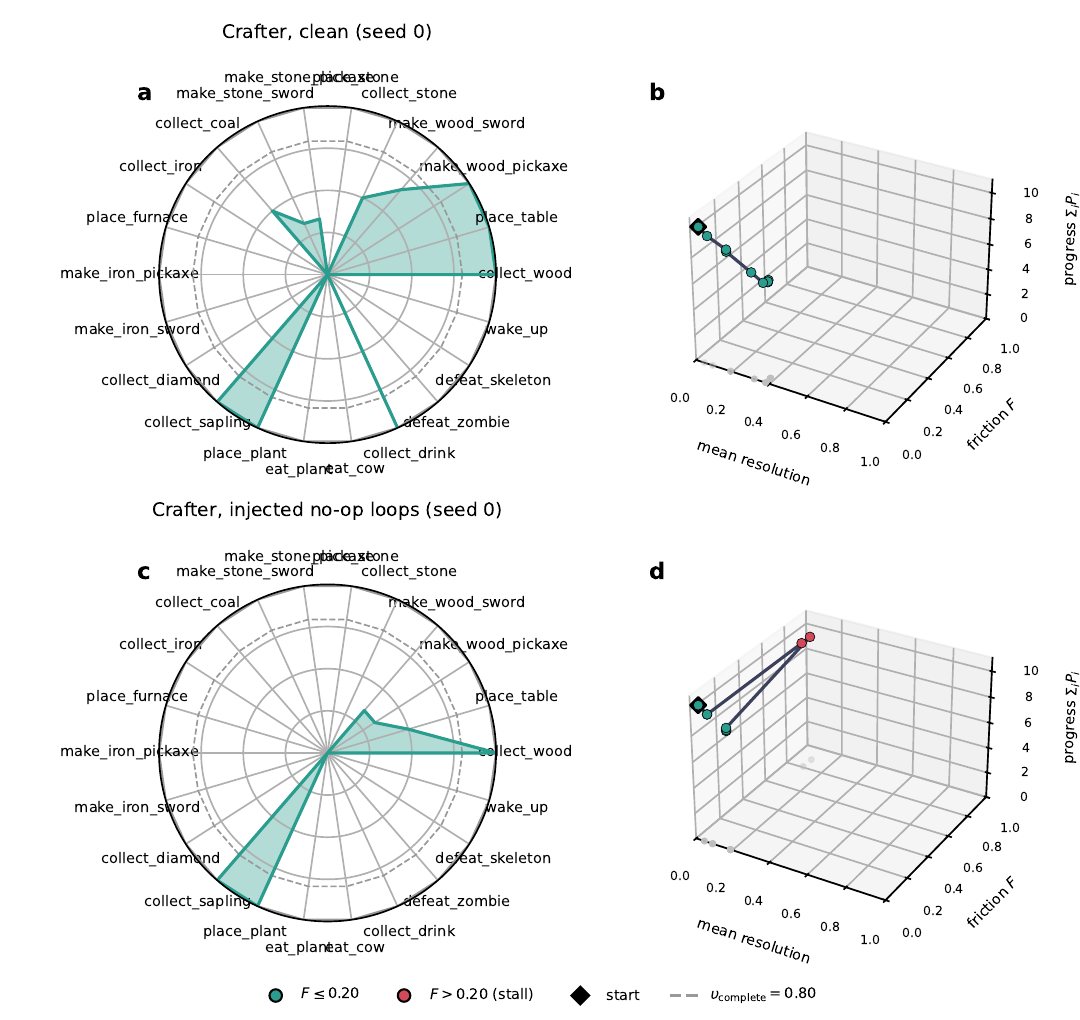}
\caption{\textbf{Crafter, seed 0.} Panels \textbf{a} and \textbf{b} show clean execution. Panels \textbf{c} and \textbf{d} show no-op loops inserted during exchanges 3 through 5 and 8 through 10.}
\label{fig:supp_crafter}
\end{figure}

\begin{figure}[!p]
\centering
\includegraphics[width=\textwidth,height=0.8\textheight,keepaspectratio]{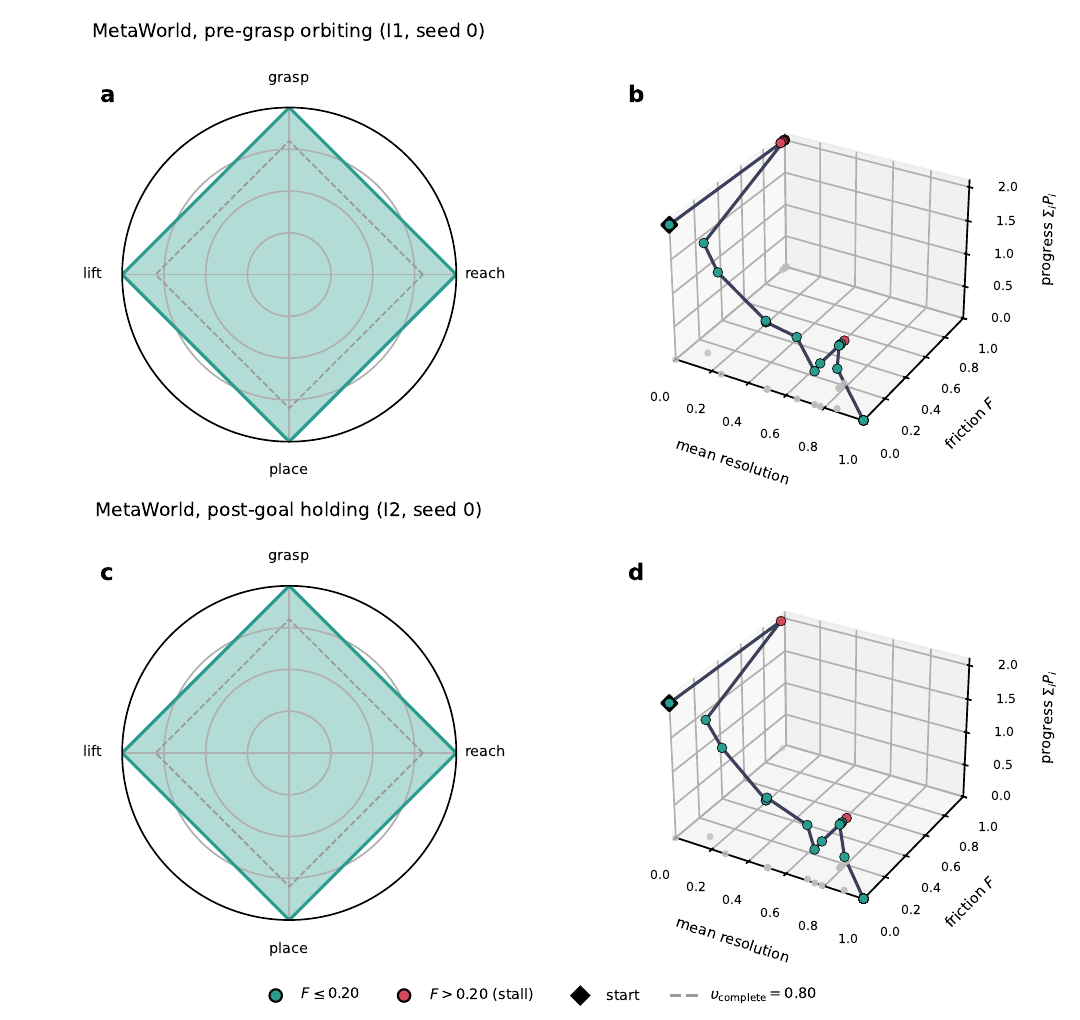}
\caption{\textbf{MetaWorld, seed 0.} Panels \textbf{a} and \textbf{b} show Condition I1. Pre-grasp orbiting occurs during exchanges 2 through 15, with friction above the diagnostic threshold on all 14 exchanges at zero attained resolution. Panels \textbf{c} and \textbf{d} show Condition I2. Goal holding occurs during exchanges 22 through 35 after task completion at exchange 17. No holding exchange exceeds the friction threshold.}
\label{fig:supp_metaworld}
\end{figure}

\begin{figure}[!p]
\centering
\includegraphics[width=\textwidth,height=0.8\textheight,keepaspectratio]{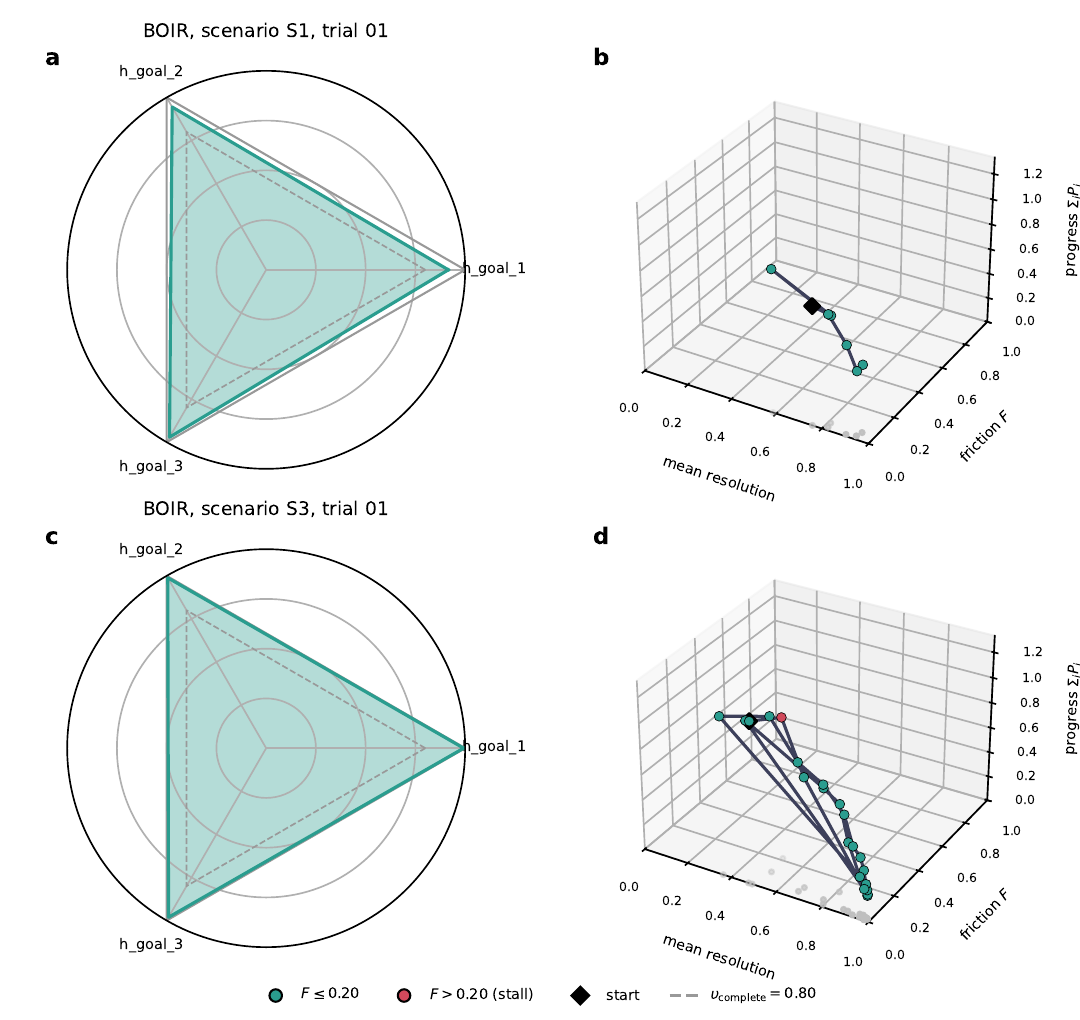}
\caption{\textbf{BOIR intent estimator, trial 01 in scenarios S1 and S3.} The task coordinates correspond to goal hypotheses. The monitor resets after each goal change. The resolution diagram represents the final epoch, and the acquisition trajectory returns towards the origin after each reset.}
\label{fig:supp_boir}
\end{figure}

\begin{figure}[!p]
\centering
\includegraphics[width=\textwidth,height=0.8\textheight,keepaspectratio]{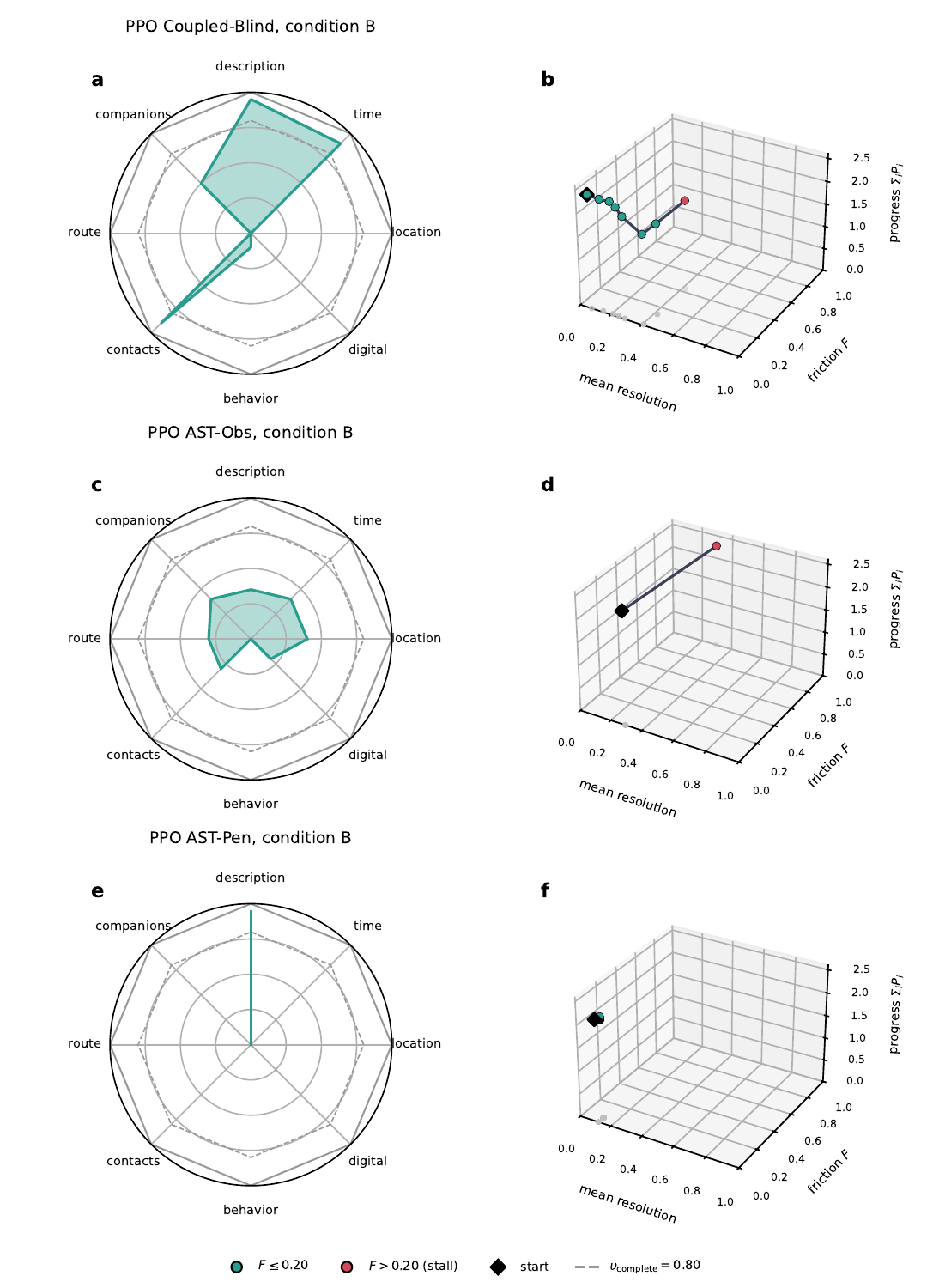}
\caption{\textbf{Closed-loop PPO episodes under Condition B, child corpus, seed 0.} The figure shows Coupled-Blind, AST-Obs and AST-Pen policies. Figure~\ref{fig:RL_viz} shows Baseline and Full-AST policies under the same condition.}
\label{fig:supp_rl}
\end{figure}

Appendix Figures~\ref{fig:supp_sar_hiker} through~\ref{fig:supp_rl} show representative trajectories generated by \texttt{experiments/figures.py} from the released data files. For Map Task, MultiWOZ and AnnoMI, the selected sessions illustrate the lowest and highest friction cases in the evaluated samples. The left-hand panels show final attained resolution for each requirement, the full-resolution boundary and the completion threshold $\upsilon_{\mathrm{complete}}=0.80$. The right-hand panels show trajectories through mean resolution, friction and total progress. Red markers indicate exchanges with $\mathrm{SI}^{\perp}>\theta_{\mathrm{SI}}=0.20$. A diamond identifies the first exchange. These plots show the acquisition state, so consecutive exchanges with identical telemetry may overlap.

\end{document}